%% file: main.tex
\documentclass{article}

\usepackage[final]{neurips_2025}

\usepackage[T1]{fontenc}    
\usepackage{url}            
\usepackage{booktabs}       
\usepackage{amsfonts}       
\usepackage{nicefrac}       
\usepackage{microtype}      
\usepackage{xcolor}         
\usepackage{caption}
\usepackage{subcaption} %
\usepackage{graphicx}
\definecolor{mydarkblue}{rgb}{0.68, 0.85, 1.0}

\usepackage[T1]{fontenc}    

\usepackage{url}            
\usepackage{booktabs}       
\usepackage{amsfonts}       
\usepackage{nicefrac}       
\usepackage{microtype}      

\usepackage{sidecap}
\usepackage{float}
\usepackage{graphicx}

\usepackage{mathrsfs}

\usepackage{amsmath}
\usepackage{amssymb}
\usepackage{amsthm}
\usepackage{mathtools}
\usepackage[mathscr]{eucal}%
\usepackage{bm}
\usepackage{bbm}
\usepackage{dsfont}
\usepackage{relsize}
\usepackage{graphics}
\usepackage{wrapfig}
\usepackage{multirow}
\usepackage{enumitem}
\usepackage{tablefootnote}
\usepackage{makecell}

\usepackage{array}
\usepackage{color, colortbl}
\definecolor{Gray}{gray}{0.9}
\definecolor{darkpastelgreen}{rgb}{0.01, 0.75, 0.24}
\definecolor{cadetgrey}{rgb}{0.57, 0.64, 0.69}
\definecolor{camel}{rgb}{0.76, 0.6, 0.42}
\definecolor{lightskyblue}{rgb}{0.53, 0.81, 0.98}
\definecolor{lightsblue}{rgb}{0.53, 0.81, 0.98}
\definecolor{lightblue}{rgb}{0.68, 0.85, 0.9}
\definecolor{softblue}{rgb}{0.85, 0.91, 0.98}
\definecolor{mygray}{gray}{0.6}
\usepackage{algorithm}
\usepackage{setspace}
\usepackage{algpseudocode}
\usepackage{pgf}
\usepackage{xinttools}
\usepackage{tikz}
\usetikzlibrary{arrows.meta}
\usepackage{textcomp}
\usetikzlibrary{calc,shapes,arrows,positioning,automata,trees}
\usepackage{placeins} 
\usepackage{tabularx}
\usepackage{graphicx}
\usepackage{subcaption} 
\usepackage{listings}

\usepackage[font=small,labelfont=bf,tableposition=top]{caption}

\DeclareCaptionLabelFormat{andtable}{#1~#2  \&  \tablename~\thetable}

\usepackage{blindtext}

\usepackage{bigdelim}
\usepackage{colortbl} 
\definecolor{mColor1}{rgb}{0.95,0.95,0.95}

\newcommand{\ours}[0]{Variational Task Vector Composition\xspace}

\usepackage{microtype}
\usepackage{graphicx}
\usepackage{adjustbox}
\usepackage{booktabs} 
\usepackage{soul}

\definecolor{mydarkblue}{rgb}{0.68, 0.85, 1.0}
\definecolor{mydarkblue2}{rgb}{0,0.08,0.45}
\definecolor{mydarkblue3}{RGB}{151,204,255}
\definecolor{cvprblue}{rgb}{0.21,0.49,0.74}

\usepackage[colorlinks=true,
    linkcolor=mydarkblue2,
    citecolor=cvprblue,
    filecolor=cvprblue,
    urlcolor=mydarkblue2]{hyperref}

\makeatletter
\renewcommand{\paragraph}{%
  \@startsection{paragraph}{4}%
  {\z@}{0em}{-1em}%
  {\normalfont\normalsize\bfseries}%
}
\makeatother    

\title{Variational Task Vector Composition}

\author{%
  \textbf{Boyuan Zhang}$^{1}$ \quad
  \textbf{Yingjun Du}$^{2}$ \quad
  \textbf{Xiantong Zhen}$^{3}$ \quad
  \textbf{Ling Shao}$^{1}$\thanks{Corresponding author. \texttt{ling.shao@ieee.org}} \\
  $^{1}$UCAS-Terminus AI Lab, University of Chinese Academy of Sciences \\
  $^{2}$AIM Lab, University of Amsterdam \\
  $^{3}$Central Research Institute, United Imaging Healthcare, Co., Ltd.
}

\begin{document}

\maketitle

\input{sections/0_abstract}
\input{sections/1_introduction}

\input{sections/2_related}
\input{sections/3_preliminary}
\input{sections/4_methods}
\input{sections/5_experiments}

\input{sections/6_conclusion}

\medskip

\newpage

\bibliographystyle{abbrv} 
\bibliography{main}

\newpage	
\appendix
\input{sections/7_appendix}
\clearpage
\input{sections/8_checklist}

\end{document}

%% file: sections/0_abstract.tex
\begin{abstract}

Task vectors capture how a model changes during fine-tuning by recording the difference between pre-trained and task-specific weights. The composition of task vectors, a key operator in task arithmetic, enables models to integrate knowledge from multiple tasks without incurring additional inference costs. In this paper, we propose variational task vector composition, where composition coefficients are taken as latent variables and estimated in a Bayesian inference framework. Unlike previous methods that operate at the task level, our framework focuses on sample-specific composition. Motivated by the observation of structural redundancy in task vectors, we introduce a Spike-and-Slab prior that promotes sparsity and preserves only the most informative components. To further address the high variance and sampling inefficiency in sparse, high-dimensional spaces, we develop a gated sampling mechanism that constructs a controllable posterior by filtering the composition coefficients based on both uncertainty and importance. This yields a more stable and interpretable variational framework by deterministically selecting reliable task components, reducing sampling variance while improving transparency and generalization. Experimental results demonstrate that our method consistently outperforms existing approaches across all datasets by selectively leveraging the most reliable and informative components in task vectors. These findings highlight the practical value of our approach, establishing a new standard for efficient and effective task vector composition.

\end{abstract}

%% file: sections/1_introduction.tex
\section{Introduction}
\label{sec: introduction}

Task vectors represent the difference between a model’s pre-trained and fine-tuned weights, capturing the changes during fine-tuning on specific tasks. Using task vectors, task arithmetic \citep{task_arith_2023} enables simple and efficient model editing. The composition of task vectors, a key operator in task arithmetic, enables models to integrate knowledge from multiple tasks without increasing additional inference costs. Task vector composition has demonstrated strong performance across a variety of domains, including computer vision \citep{finding_visual_2024}, natural language processing \citep{function_vectors_llm_2024, generalization_analysis_2025, in_context_task_vectors_2023}, and multimodal learning \citep{multimodal_task_vector_2024, vlm_task_vector_2025}. This approach provides a practical solution for knowledge integration and model editing \citep{model_merging_2024, knowledge_editing_survey_2024, comprehensive_knowledge_editing_2024, knowledge_conflicts_2024}.

In recent years, the composition of task vectors has been widely studied to improve efficiency and controllability \citep{composing_arithmetic_2023}. Some works focus on improving task vector representations by redesigning spaces and developing parameterization techniques to better capture the relationships between tasks \citep{tangent_space_2023, task_singular_vectors_2024}. Others extend the functional scope of task vectors to reduce computational overhead \citep{function_vectors_llm_2024, finding_visual_2024, badtv_2025}. Advances in composition methods also enable more effective integration of knowledge across multiple tasks, enhancing both the controllability and interpretability of models \citep{knowledge_composition_2024, dynamic_grouping_2025}. While recent works have made notable progress, three key challenges remain under-explored. Existing approaches typically rely on deterministic composition schemes, lacking mechanisms to quantify uncertainty. Moreover, most methods operate at the task level, which limits their ability to adapt to sample-level variability. Additionally, task vector spaces often exhibit redundant structure, which may reduce efficiency in both inference and storage.

In this paper, we propose a novel approach for the variational composition of task vectors. Our main contributions are as follows: \textit{(i)} We cast task vector composition as a variational inference problem and introduce an amortized inference network to model sample-specific posteriors over composition coefficients. This formulation enables efficient integration of domain-specific knowledge at the sample level and overcomes the limitations of conventional task-level composition.
\textit{(ii)} To address the structural redundancy in task vector spaces, we introduce a Spike-and-Slab prior that promotes sparsity by modeling the variational posterior as a mixture of zero-valued \textit{spikes} and Gaussian \textit{slabs}. This structured prior guides the variational posterior to focus on essential task components, enhancing interpretability and improving the efficiency of sample-specific composition. 
\textit{(iii)} To address the computational cost and instability of traditional sampling-based inference, we develop a controllable posterior via gated sampling, which deterministically filters composition coefficients based on their estimated uncertainty and importance.
By deterministically selecting task coefficients based on uncertainty and importance, the proposed framework avoids the high variance of stochastic sampling, highlights the components that contribute to adaptation, and ensures consistent inference, making it more stable, interpretable, and reliable.

We evaluate our framework through extensive experiments on a variety of benchmark tasks. The experimental results show that our approach consistently outperforms existing methods for task vector composition and achieves better performance in model editing applications. In addition, we conduct in-depth analyses of our three main contributions: the composition of task vectors under variational inference, the introduction of Spike-and-Slab priors, and the development of a gated sampling process. These analyses show the advantages of our approach in integrating sample-level knowledge, eliminating redundant information, and improving inference stability.

%% file: sections/2_related.tex
\section{Related work}
\vspace{-0.5em}

\noindent\textbf{Task vector composition and task arithmetic.} A task vector represents the difference between pre-trained and fine-tuned model weights, which has been widely used for model editing and multi-task learning. Ilharco et al. \citep{task_arith_2023} first demonstrated that arithmetic operations in the weight space can enhance multi-task performance and enable task forgetting. Ortiz-Jiménez et al. \citep{tangent_space_2023} further established the theoretical basis for task arithmetic by introducing the Neural Tangent Kernel perspective. Building on this, Zhang et al. \citep{knowledge_composition_2024} proposed the aTLAS framework, which uses anisotropic scaling to enable more precise parameter adjustment and reduce computational overhead. In language models, research on task vectors and function vectors has shown that neural networks can learn compact and transferable representations with supporting compositional knowledge transfer and interpretable causal reasoning \citep{finding_visual_2024, function_vectors_llm_2024}. Recent theoretical work has identified key factors for effective task arithmetic in nonlinear models, such as task independence, consistency, and optimal coefficient selection~\citep{generalization_analysis_2025}. Despite these advances, most prior works rely on deterministic composition at the task level, overlooking the need for sample-specific adaptation and lacking mechanisms to model uncertainty in the coefficient selection.
We address these gaps by casting task vector composition as variational inference, with sparse priors and gated posterior construction for improved efficiency and reliability.

\noindent\textbf{Applications of Spike-and-Slab priors.} Spike-and-Slab priors are widely used in Bayesian sparse modeling to distinguish between important and unimportant variables \citep{spike_slab_posterior_2025, spike_slab_lasso_2021, spike_slab_discovery_2021}. The foundational work by Mitchell et al. \citep{bayesian_variable_linear_1988} introduced mixture distributions for variable selection, enabling effective sparsity. Ishwaran et al. \citep{spike_slab_selection_2005} extended this approach by estimating variable importance via MCMC sampling. Traditional Spike-and-Slab methods face scalability challenges on large datasets, which has prompted a shift toward variational inference for greater efficiency. For example, Titsias et al. \citep{spike_slab_vi_2011} developed a variational Spike-and-Slab method for multi-task learning, promoting shared sparse structure across tasks. Ray et al. \citep{spike_slab_logit_2020} designed a variational Bayesian approach for high-dimensional sparse logistic regression with strong theoretical guarantees. In deep learning, Goodfellow et al. \citep{large_spike_slab_2012} applied Spike-and-Slab priors to unsupervised image classification. More recent studies have proposed structured Spike-and-Slab priors to model spatial correlations, comparing various structural forms and improving modeling flexibility \citep{spike_slab_structured_2012, bayesian_spike_slab_2014, dynamic_spike_slab_2021, comparing_spike_slab_2011, fast_spike_slab_2022}. These advances have broadened the application of Spike-and-Slab priors in modern machine learning, but practical integration with deep models remains a challenge, especially for large-scale or sample-specific sparse inference. In this work, we extend Spike-and-Slab priors to the context of task vector composition in deep models, where we model the coefficient space as a structured mixture prior. By integrating this prior into a variational framework and leveraging amortized inference, we achieve scalable, sample-aware sparsity that is both interpretable and effective. 

%% file: sections/3_preliminary.tex
\section{Preliminary}
\vspace{-0.5em}

In transfer learning and multi-task learning, task-specific knowledge is often captured by modeling changes in network parameters, including models based on CLIP \citep{clip_2021}, GPT-2 \citep{gpt2_2019}, and T5 \citep{t5_2020}. In this paper, we focus particularly on CLIP, as it offers powerful multimodal representations and aligns with experimental settings used in previous work.

Formally, denote the CLIP image encoder as a function $f: \mathcal{X} \times \Theta \rightarrow \mathcal{Z}$, where for an input image $x \in \mathcal{X}$ and parameters $\theta \in \Theta$, the output $z = f(x; \theta)$ represents the learned latent representation for the input image. Given the pre-trained parameters $\Theta_0$ and the fine-tuned parameters $\Theta_t$ in a downstream task $t$, the corresponding task vector is defined as $\tau_t = \Theta_t - \Theta_0$. In practice, we follow standard procedures by fine-tuning only the image encoder while keeping the text encoder fixed, which helps maintain feature alignment across tasks. 

Task vectors can be composed through various arithmetic operations, and the most common operation is task addition. In task addition, given a task vector pool $\mathcal{T}$ with $N$ task vectors $\{\tau_i\}_{i=1}^N$ fine-tuned on different tasks, with a global learnable scaling factor $\lambda$, a unified model can be constructed as: $\Theta_{\text{new}} = \Theta_0 + \lambda \cdot \sum_{i=1}^N \tau_i$. This form of composition has shown strong practical value, enabling effective integration of knowledge from multiple tasks into a single model and demonstrating superior transfer capabilities in image classification and transfer learning benchmarks.

Task vectors can also be decomposed into block-wise representations. Based on functional partitioning, aTLAS \citep{knowledge_composition_2024} divides each task vector $\tau_i$ into $M$ parameter blocks $\{\tau_i^j\}_{j=1}^M$, and assigns an independent scaling coefficient $\Lambda_i^j$ to each block. Such modular decomposition also enables the exploration of task relevance at different network layers or functional units, which can be further exploited in downstream analysis. The composed model parameters can then be written as:

\begin{equation}
\Theta = \Theta_0 + \sum_{i=1}^N\sum_{j=1}^M \Lambda_i^j \cdot \tau_i^j.
\label{task vector composition}
\end{equation}

This block-wise formulation enables finer-grained knowledge integration and provides a foundation for efficient multi-task learning and task transfer. While most existing methods perform task vector composition at the task level, they fail to capture fine-grained differences between each sample. To address this limitation, we propose modeling task vector composition at the sample level. This finer-grained approach allows the model to adapt to input variability and enhances performance in heterogeneous environments.

%% file: sections/4_methods.tex
\section{Methodology}
\vspace{-0.5em}

This section introduces our proposed framework for the variational composition of task vectors. In Section~\ref{subsec:Variational Composition of Task Vectors}, we reformulate the composition problem as a variational inference process and introduce an amortized inference network to model sample-specific posteriors over composition coefficients. To address the structural redundancy in task vector spaces, we then introduce a Spike-and-Slab prior to promote sparsity by modeling the variational posterior as a mixture distribution in section~\ref{subsec:Spike-and-Slab Priors for Sparse Representations}. We also propose a controllable posterior with gated sampling that automatically selects informative task vectors in section~\ref{subsec:Controllable Posterior with Gated Sampling}, which enhances the model's generalization capability and robustness.

\subsection{Variational Composition of Task Vectors}
\label{subsec:Variational Composition of Task Vectors}
Conventional task vector composition is typically performed as a direct and deterministic modification of model weights, lacking any mechanism for uncertainty quantification. As a result, such approaches are less capable of adapting to sample-specific variations and cannot estimate uncertainty in the composition process. In our variational inference framework, the composition coefficients 
\begin{wrapfigure}{r}{0.3\textwidth}
    \centering
    \vspace{-2mm}
    \includegraphics[width=0.23\textwidth]{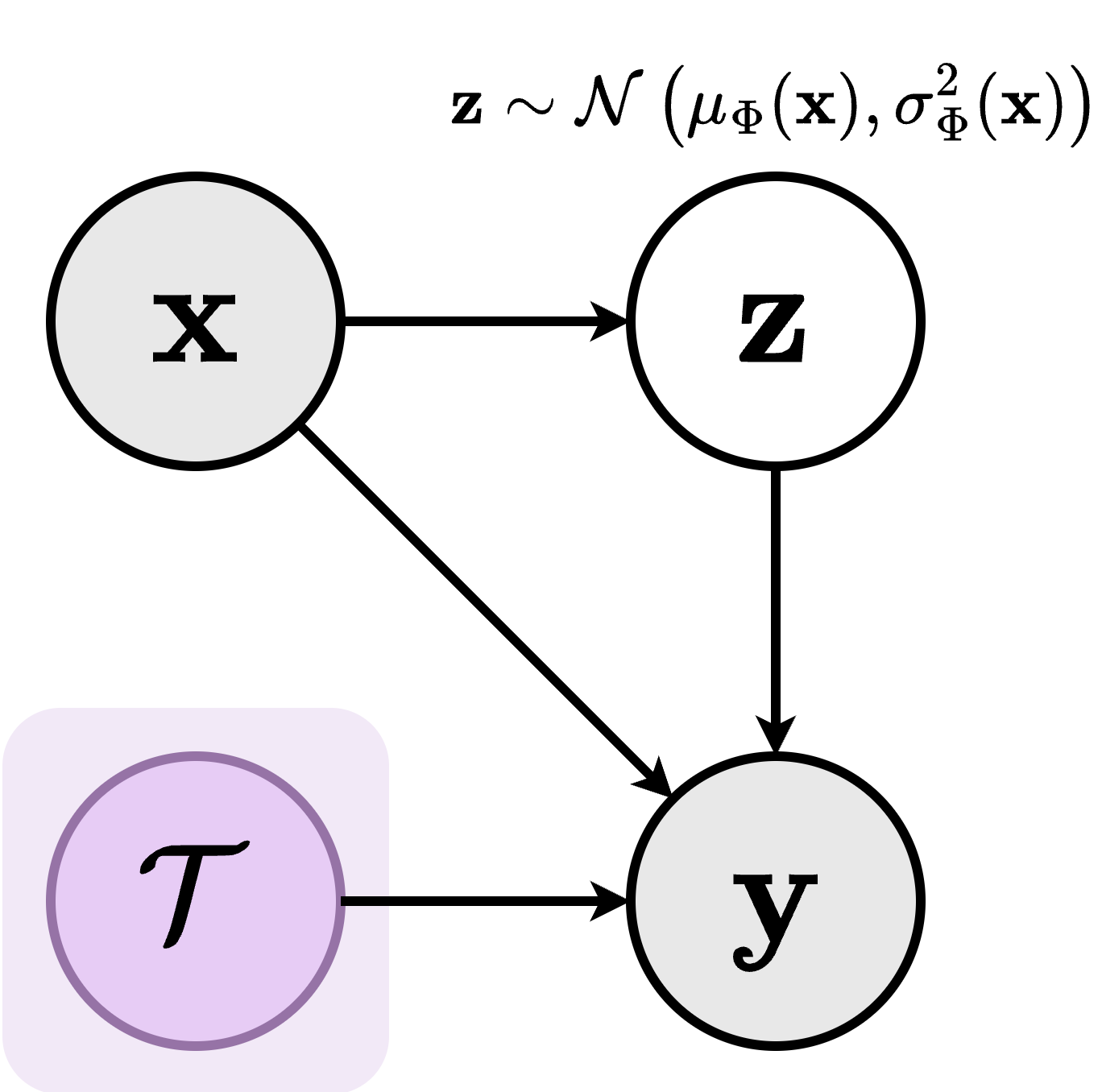}
    \caption{\textbf{Variational Composition of Task Vectors.} }
    \label{fig:task_vector_VI}
    \vspace{-1mm}
\end{wrapfigure}
$\mathbf{z}$ are treated as latent variables, while the input $\mathbf{x}$ and labels $\mathbf{y}$ are observed. Figure~\ref{fig:task_vector_VI} illustrates the probabilistic dependencies. According to Bayesian theory, our objective is to compute the posterior distribution $p(\mathbf{z}|\mathbf{x},\mathbf{y})$. Since the posterior is typically difficult to compute directly, variational inference introduces a parameterizable approximate posterior distribution $q(\mathbf{z}|\mathbf{x})$, leading to the evidence lower bound (ELBO):
\begin{equation}
    \log p(\mathbf{y}|\mathbf{x}) \geq \mathbb{E}_{q(\mathbf{z}|\mathbf{x})}[\log p(\mathbf{y}|\mathbf{x},\mathbf{z})] - D_{\mathbf{KL}}(q(\mathbf{z}|\mathbf{x})||p(\mathbf{z})).
\label{Variational ELBO}
\end{equation}
Here, the first term represents the expected log-likelihood, which promotes the model to make accurate predictions, while the second term is the KL divergence, which regularizes the posterior towards the prior to prevent overfitting. This framework naturally recasts the composition problem as an inference task, allowing us to incorporate flexible prior knowledge, such as sparsity or structural constraints, into the composition process. The detailed mathematical derivation is provided in Appendix~\ref{sec:appendixA.1}.

To efficiently realize sample-level variational inference, we employ amortized inference by designing a neural network with parameters $\Phi$ to parameterize the posterior distribution: $q_{\Phi}(\mathbf{z}|\mathbf{x}) = \mathcal{N}(\mathbf{z};\mu_{\Phi}(\mathbf{x}),\sigma_{\Phi}(\mathbf{x}))$. This network takes samples $\mathbf{x}$ as inputs and predicts the parameters of the composition coefficient distribution $\mathbf{z}$. Amortized inference improves both efficiency and scalability by sharing parameters across samples.

In summary, sample-level composition allows the model to adapt composition coefficients to individual inputs, offering finer control over task integration and leading to improved generalization in heterogeneous settings. This represents a fundamental difference from task-level approaches. However, conventional variational inference typically uses Gaussian priors, which are unable to capture or address structural redundancy in task vector spaces. This can lead to inefficient composition and increased risk of overfitting.

\subsection{Spike-and-Slab Priors for Sparse Representations}
\label{subsec:Spike-and-Slab Priors for Sparse Representations}
\begin{figure}[t]
    \centering
    \begin{subfigure}[t]{0.48\textwidth}
        \centering
        \includegraphics[width=\textwidth]{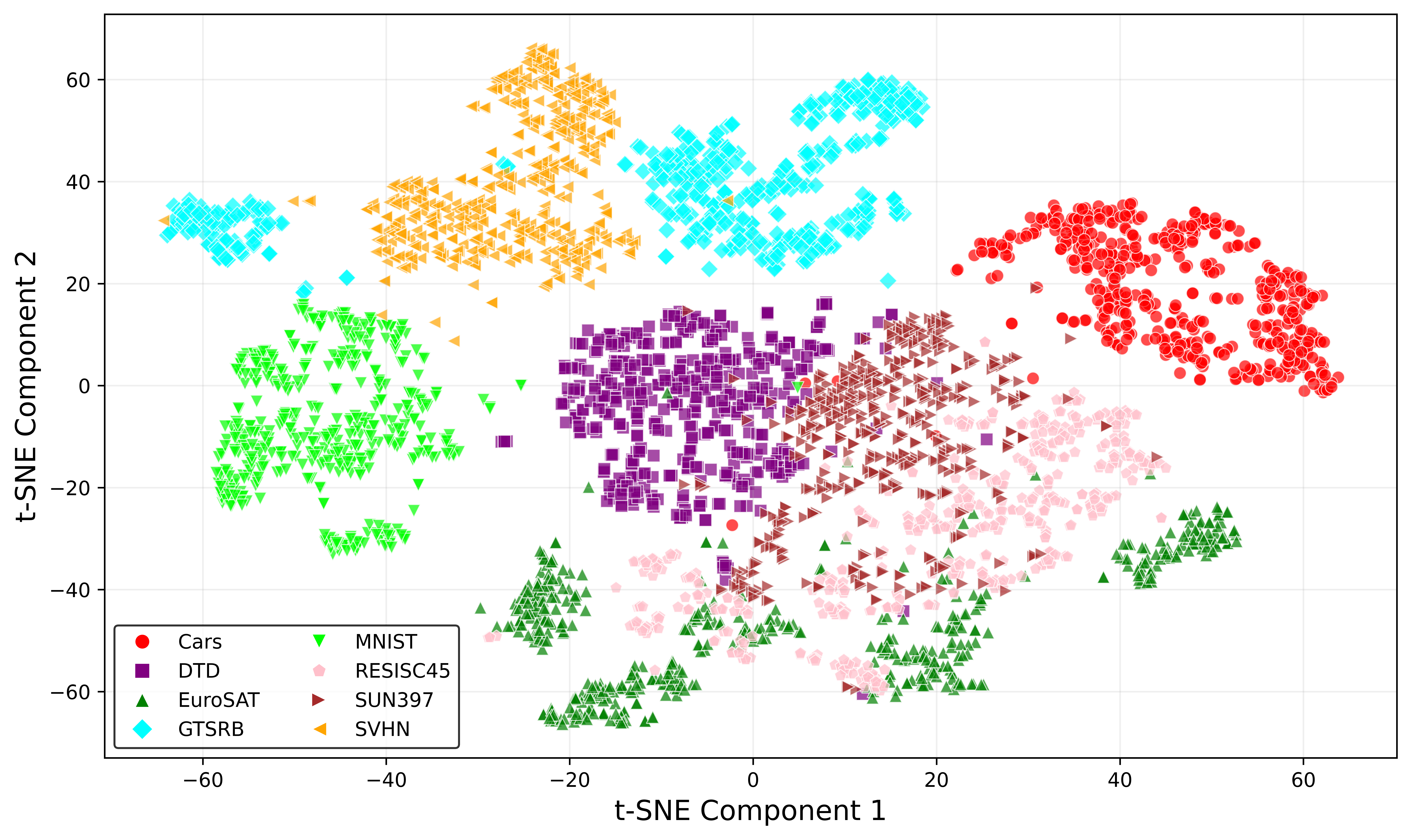}
        \vspace{-5mm}
        \caption*{(a) t-SNE clustering of task representations.}
        \label{fig:tsne_clustering}
    \end{subfigure}
    \begin{subfigure}[t]{0.48\textwidth}
        \centering
        \includegraphics[width=\textwidth]{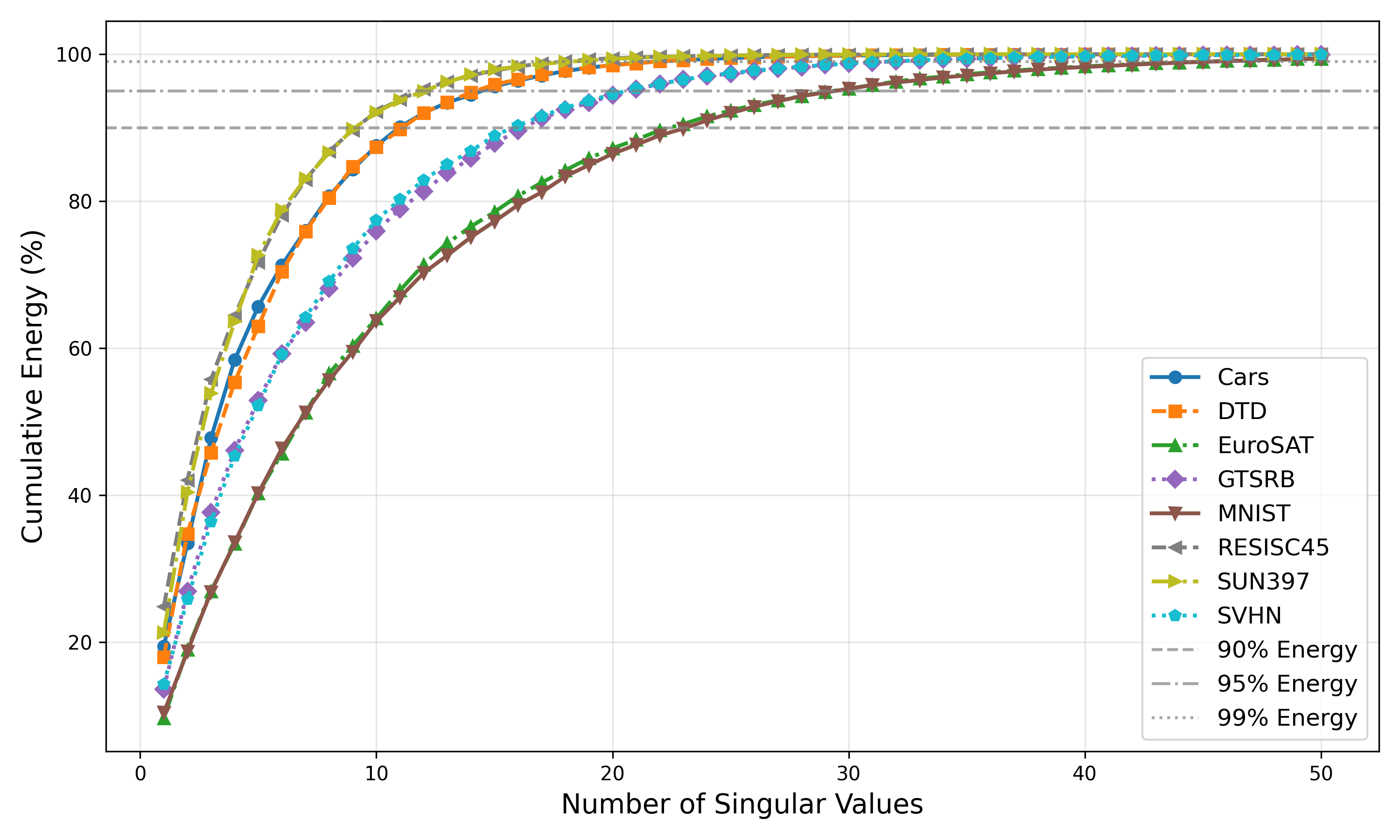}
        \vspace{-5mm}
        \caption*{(b) Cumulative singular value energy distribution.}
        \label{fig:svd_energy}
    \end{subfigure}
    \hfill
    \caption{
        \textbf{Redundancy and structural analysis of task vectors.} (a) t-SNE visualization shows that task representations form distinct clusters corresponding to each dataset, but there are notable overlap regions, indicating partial coupling of task features across domains. (b) Cumulative singular value energy plots reveal that fewer than 40 principal components account for over 95\% of the variance across all eight datasets. These analyses reveal substantial structural redundancy in task vector representations.
    }
    \label{fig:redundancy_analysis}
    \vspace{-5mm}
\end{figure}

The standard Gaussian prior is widely used, but it fails to capture the structural properties inherent in high-dimensional task vector spaces. As illustrated in Fig.~\ref{fig:redundancy_analysis}, we conduct a thorough redundancy analysis of task vector representations. Using t-SNE and singular value decomposition, we find significant redundancy in task vectors, which leads to inefficient integration and a higher risk of overfitting. For instance, over 95\% of the variance can be explained by fewer than 40 principal components across all datasets, highlighting substantial compressibility. Moreover, when input samples are located near the boundaries of multiple task domains, Gaussian priors are unable to capture the multimodal nature of the underlying composition, thereby limiting model performance on heterogeneous data.

Our variational composition framework offers a flexible way of incorporating prior knowledge to address the issue of structural redundancy in task vector representations. We introduce a Spike-and-Slab prior, which effectively promotes structural sparsity in the compositions of task vectors. The Spike-and-Slab prior is defined as a mixture distribution:
\begin{equation}
    p(\mathbf{z}) = (1 - \pi)\,\delta_0(\mathbf{z}) + \pi\,\mathcal{N}(\mathbf{z};0,\sigma^2),
\label{Spike-and-Slab prior}
\end{equation}
where $\delta_0(\mathbf{z})$ denotes the Dirac delta distribution at zero, $\mathcal{N}(\mathbf{z}; 0, \sigma^2)$ denotes a Gaussian distribution with zero mean and variance $\sigma^2$, and $\pi$ is the mixture coefficient that controls the expected sparsity of the coefficients. The Spike-and-Slab prior allocates probability mass between a point mass at zero (the \textit{spike}) and a continuous Gaussian distribution (the \textit{slab}). This ensures that coefficients are either exactly zero or drawn from the Gaussian, thereby enabling explicit and interpretable structure selection. By promoting coefficient sparsity, the Spike-and-Slab prior automatically identifies and retains the most important task vector components while effectively eliminating irrelevant ones. This property not only improves model efficiency but also enhances interpretability, allowing us to clearly determine which task knowledge is relevant to each input sample.

Under the Spike-and-Slab prior, we adopt a factorized variational posterior:
\begin{equation}
q_\Phi(\mathbf{z}|\mathbf{x}) = \prod_i\prod_j q_\Phi(\omega_i^j|\mathbf{x}),q_\Phi(\mathbf{z}_i^j|\omega_i^j,\mathbf{x}),
\label{variational posterior}
\end{equation}
Specifically, for each composition coefficient $\mathbf{z}_i^j$, we introduce a corresponding binary indicator variable $\omega_i^j$. The coefficient is retained when $\omega_i^j=1$ and set to zero when $\omega_i^j=0$.

By performing variational inference, we obtain the following ELBO objective:

\begin{equation}
    \mathcal{L}_{\mathbf{ELBO}} = \mathbb{E}_{q_\Phi(\mathbf{z},\omega|\mathbf{x})}[\log p(\mathbf{y}|\mathbf{x},\mathbf{z},\omega)] - D_{\bf{KL}}(q_\Phi(\mathbf{z},\omega | \mathbf{x})||p(\mathbf{z},\omega)),
\label{Spike-and-Slab ELBO}
\end{equation}

To optimize this framework, we employ amortized inference with Monte Carlo sampling. We design a neural network parameterized by $\Phi$ that inputs sample $\mathbf{x}$ and outputs three sets of parameters: inclusion probabilities $\pi_i^j(\mathbf{x})$, weight means $\mu_{ij}(\mathbf{x})$, and log variances $\log\sigma_{ij}^2(\mathbf{x})$. During training, we first sample binary inclusion variables $\omega_i^j \sim \mathrm{Bernoulli}(\pi_i^j(\mathbf{x}))$, then sample weights $\mathbf{z}_i^j \sim \mathcal{N}(\mu_{ij}(\mathbf{x}), \sigma_{ij}^2(\mathbf{x}))$, and finally combine them to obtain sparse composition coefficients $\tilde{\mathbf{z}}_i^j = \omega_i^j \cdot \mathbf{z}_i^j$. The complete derivation of the variational objective with Spike-and-Slab prior is available in Appendix~\ref{sec:appendixA.2}.

However, in high-dimensional settings, Monte Carlo-based variational inference often suffers from unstable sparsity patterns and unreliable results \cite{rethinking_VI_2022, challenges_high_dim_2021}. This limitation motivates the development of a more stable and controllable posterior, which we describe in the following section.

\subsection{Controllable Posterior with Gated Sampling}
\label{subsec:Controllable Posterior with Gated Sampling}

\begin{figure}[t]
    \centering
    \includegraphics[width=1\textwidth]{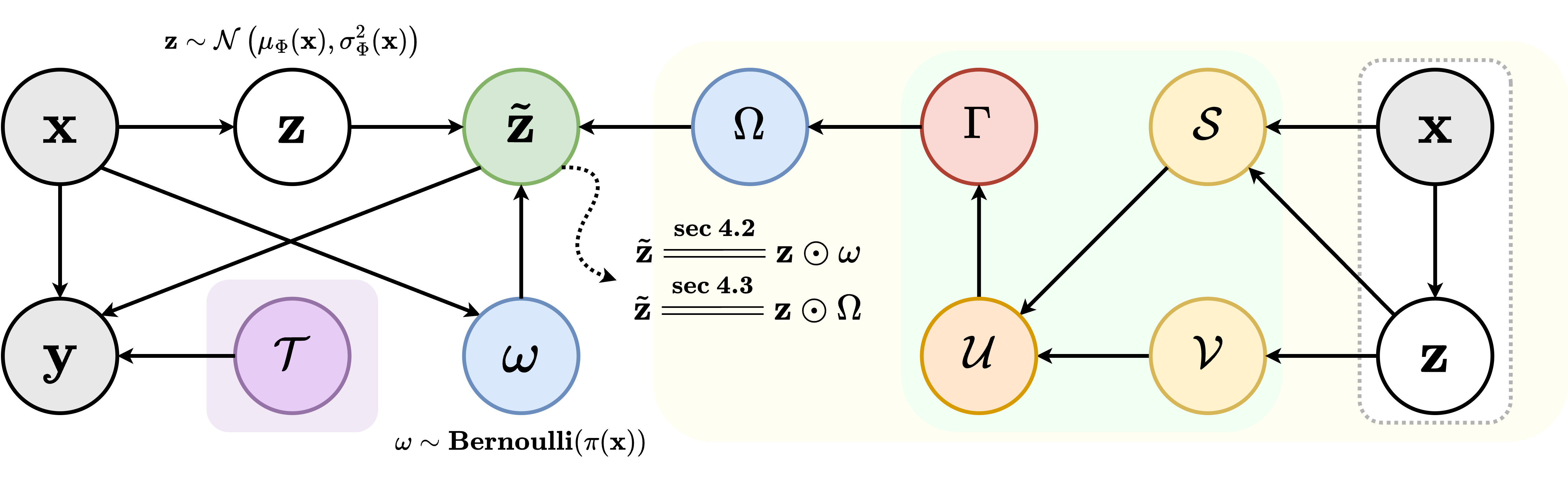}
    \caption{\textbf{Variational Composition of Task Vectors with Spike-and-Slab Prior}. The framework illustrates two inference approaches based on the Spike-and-Slab prior. On the left, the sampling-based variational posterior process generates latent variables $\mathbf{z}$ from input features $\mathbf{x}$ via $\mathcal{N}(\mu_\Phi(\mathbf{x}), \sigma_\Phi^2(\mathbf{x}))$, producing sparse representations $\tilde{\mathbf{z}} = \mathbf{z} \odot \omega$ through element-wise multiplication with binary indicators $\omega$ sampled from $\mathrm{Bernoulli}(\pi(\mathbf{x}))$. On the right, the deterministic gated sampling process computes uncertainty $\mathcal{U}$ from gradient sensitivity $\mathcal{S}$ and distributional deviation $\mathcal{V}$, uses an adaptive threshold $\Gamma$ to generate continuous gating variables $\Omega$, and produces sparse representations $\tilde{\mathbf{z}} = \mathbf{z} \odot \Omega$. This framework demonstrates the transition from random binary selection to deterministic continuous gated in the inference process.}
    \label{fig:prior_posterior_graph}
    \vspace{-5mm}
\end{figure}

To address the instability and high variance of Monte Carlo sampling in high dimensions, we propose a gated sampling mechanism for controllable posterior. Instead of sampling binary variables from the conventional Bernoulli distribution with probability $\pi(\mathbf{x})$, our method employs a deterministic, continuous gating function to select composition coefficients. 
The continuous gating function replaces discrete sampling with 
differentiable alternative, enabling gradient-based training while preserving the structure of variational inference.
Figure~\ref{fig:prior_posterior_graph} illustrates both stochastic and deterministic inference processes, highlighting how our method transitions from random Bernoulli sampling to uncertainty-aware, soft-gated posterior construction.

We begin by constructing an uncertainty estimate $\mathcal{U}$, which integrates two components: gradient sensitivity and distributional deviation. Gradient sensitivity $\mathcal{S}$ quantifies how responsive a coefficient is to small perturbations in the input, while distributional deviation $\mathcal{V}$ measures the extent to which a coefficient deviates from the batch mean, where $\mu_B$ and $\sigma_B$ denote the mean and standard deviation of the coefficient within the batch, respectively. The overall uncertainty measure is formulated as a weighted combination:
\begin{equation}
\mathcal{U} = \eta \cdot \mathcal{S} + (1 - \eta) \cdot \mathcal{V} = \eta \cdot |\nabla_{\mathbf{x}} \mathbf{z}(\mathbf{x})|_2 + (1-\eta) \cdot \left|\frac{\mathbf{z}(\mathbf{x})-\mu_B}{\sigma_B}\right|,
\label{uncertainty measure}
\end{equation}
where $\eta$ is a balancing coefficient. Intuitively, coefficients that are highly sensitive to input changes or show large fluctuations among similar samples are deemed unreliable and should be suppressed, while stable and significant coefficients should be preserved.

We then define an adaptive gating function:
\begin{equation}
    \mathcal{G}\left(\mathbf{z}, \mathcal{U}; \Psi\right) = \mathbf{z} \cdot \mathbb{I}(|\mathbf{z}| \geq \Gamma(\mathcal{U};\Psi)),
\label{gated function}
\end{equation}
where the threshold function $\Gamma(\mathcal{U};\Psi)$ varies with the uncertainty metric $\mathcal{U}$, creating an adaptive selection mechanism as follows:
\begin{equation}
    \Gamma(\mathcal{U};\Psi)=\psi_1 \cdot (1+\psi_2 \cdot \mathcal{U}),
\label{selection mechanism}
\end{equation}
where $\Psi=\{\psi_1, \psi_2\}$ are learnable parameters, $\psi_1$ is a base threshold, and $\psi_2$ controls sensitivity to uncertainty. Under this mechanism, coefficients with high uncertainty are subjected to higher thresholds and thus are less likely to be selected, while those with low uncertainty face lower thresholds, increasing their likelihood of being retained. This adaptive selection ensures that more reliable and informative components are preferentially integrated into the final model composition.

To enable end-to-end differentiable training, we approximate the hard-gated function with a soft-gated function, thereby replacing the Monte Carlo sampling methods described in Section~\ref{subsec:Spike-and-Slab Priors for Sparse Representations} with a deterministic point estimation approach:
\begin{equation}
    \Omega(\mathbf{z},\mathcal{U}) = \sigma\left(\frac{|\mathbf{z}| - \Gamma(\mathcal{U};\Psi)}{\rho}\right),
\label{soft gated}
\end{equation}
where $\sigma(\cdot)$ denotes the sigmoid function and $\rho$ is a temperature hyperparameter. During training, the soft-gated function allows for gradient-based optimization, while at inference time we switch to hard thresholding to ensure sparse and interpretable coefficient selection. Based on this design, each coefficient’s variational posterior can be approximated as a mixture of Spike-and-Slab priors:
\begin{equation}
    q(\mathbf{z}|\mathbf{x}) = [1 - \Omega(\mathbf{z},\mathcal{U})]\cdot\delta_0(\mathbf{z}) + \Omega(\mathbf{z},\mathcal{U})\cdot\mathcal{N}(\mathbf{z};\mu_\Phi(\mathbf{x}),\sigma_\Phi^2(\mathbf{x})).
\label{gated Spike-and-Slab posterior}
\end{equation}
Here, $\Omega$ replaces the binary indicator variable $\omega$ for deterministic gating. Furthermore, we parameterize the variational posterior using a Dirac delta distribution $\delta_{\mathbf{z} = f_\Phi(\mathbf{x})}$ to represent coefficients $\mathbf{z}$ as deterministic functions of input $\mathbf{x}$, while establishing deterministic relationships between gated variables $\Omega$ and coefficients $\mathbf{z}$ through $q_\Psi(\Omega|\mathbf{z})$, as follows:
\begin{equation}
    q_{\Phi,\Psi}(\Omega, \mathbf{z}|\mathbf{x}) = q_\Psi(\Omega|\mathbf{z}) \delta_{\mathbf{z} = f_\Phi(\mathbf{x})}.
\label{gated posterior}
\end{equation}
This parameterization enables the selection of deterministic coefficients within the variational inference framework, while explicitly modeling uncertainty.

In addition to the primary objective of the ELBO, we introduce auxiliary regularization losses $\mathcal{L}_{\text{reg}}$ to further stabilize training and encourage exploration. These include a boundary loss, penalizing coefficients near the threshold to encourage more decisive selection decisions, and an exploration loss, promoting threshold-related parameters to search in a broader space to avoid local optima. Detailed definitions of these regularization terms and hyperparameter settings are provided in Appendix~\ref{sec:appendixB}.

In summary, the final training objective can be expressed as:
\begin{equation}
    \mathop{\arg\min}\limits_{\Phi, \Psi} \frac{1}{|\mathcal{D}_t|} \sum_{s=1}^{|\mathcal{D}_t|}\mathcal{L}\left(f\left(\mathbf{x}_s; \Theta_0 + \mathcal{G}\left (\mathbf{z}(\mathbf{x_s};\Phi), \mathcal{U};\Psi\right)\mathcal{T}\right),\mathbf{y_s}\right) + \lambda\mathcal{L}_{\text{reg}}(\mathbf{z}, \mathcal{U}; \Psi),
\label{final training objective}
\end{equation}
where $\lambda$ is the balance parameter. Our final training objective comprises two components: the primary prediction loss based on the gated sampling, and a regularization term that stabilizes the optimization process while enhancing the model's generalization capacity. By modeling coefficient uncertainty explicitly, our method enhances generalization and interpretability. Overall, this uncertainty-guided, gated composition yields a more robust and interpretable mechanism for knowledge integration across diverse tasks.


%% file: sections/5_experiments.tex
\section{Experiments}
\label{sec: experiment}
\subsection{Experiments Setup}
\paragraph{Datasets.}
Following previous practice, we conduct experiments on eight image classification datasets: Cars \citep{standford_cars_2013}, DTD \citep{dtd_2014}, EuroSAT \citep{eurosat_2018}, GTSRB \citep{gtsrb_2011}, MNIST \citep{mnist_1998}, RESISC45 \citep{resisc45_2017}, SUN397 \citep{sun397_2016} and SVHN \citep{svhn_2011}. We select these datasets to evaluate the effectiveness and scalability of our approach on diverse visual and task scenarios.

\paragraph{Implementation Details.}
To ensure a fair comparison, we employ three pre-trained Vision Transformer architectures \citep{vit_2021} as feature extractors: $\text{ViT-B/16}$, $\text{ViT-B/32}$ and $\text{ViT-L/14}$, all pre-trained under the CLIP framework \citep{clip_2021}. The experiments use AdamW \citep{adamw_2019} Optimizer with a batch size of $128$, an initial learning rate of $0.0005$, and a weight decay of $0.01$, training for $20$ epochs in the schedule of the cosine learning rate. For the proposed controllable posterior, we set the base threshold ($\psi_1$) at $0.05$, the sensitivity parameter ($\psi_2$) at $1.0$, and the uncertainty regularization coefficient at $0.01$. To improve computational efficiency, we precompute feature representations for all datasets and apply data augmentation to generate multiple feature versions, thereby constructing a more diverse training set. All our experiments are conducted on eight NVIDIA A40 GPUs with 48GB of memory each. The details and hyperparameter settings are included in Appendix~\ref{sec:appendixB}.

\paragraph{Evaluation Metrics.}
Following \cite{knowledge_composition_2024}, we use absolute accuracy as our evaluation metric. In addition, we calculate the gated ratio to verify the effectiveness of our controllable posterior.

\subsection{Results}
\paragraph{Benefit of sample-specific task vectors.}
Sample-specific task vector composition consistently outperforms task-level baselines, as shown in Table~\ref{tab:deterministic_vs_VI}. The improvements are especially clear on $\text{Cars}$, $\text{SVHN}$, and $\text{GTSRB}$ datasets. These datasets show greater diversity within the sample and differences between domains. $\text{SVHN}$ contains data features that differ significantly from pre-training source data. $\text{GTSRB}$ includes various types of traffic signs. Cars features different automobile models and viewing angles. These characteristics make unified task-level vectors insufficient for capturing sample-specific differences. Our approach calculates combination coefficients for each input sample, enabling more accurate and adaptive knowledge integration. As a result, improvements are greatest on datasets with high variability and notable domain gaps. This confirms the effectiveness of our sample-specific task vectors when handling complex data scenarios. All results are based on features encoded by the $\text{ViT-B/32}$ model. More detailed results are provided in Appendix~\ref{sec:appendixC.1}.

\paragraph{Benefit of variational composition of task vectors.}
Table~\ref{tab:deterministic_vs_VI} shows that the variational inference framework enhances the overall performance of the model. This is particularly evident in datasets with stronger differences between domains, such as $\text{EuroSAT}$, $\text{MNIST}$, and $\text{SVHN}$. This improvement is attributed to explicit uncertainty modeling of task vector coefficients. This provides better robustness when processing cross-domain data variations. The KL divergence regularization effectively reduces overfitting risk, improving model generalization. Deterministic methods perform slightly better on $\text{DTD}$, likely due to the regularity of its texture features. In this case, deterministic inference adequately captures the necessary information. The uncertainty of the additional parameter from variational inference may introduce minor interference. The results validate that our variational inference framework is particularly effective for heterogeneous and cross-domain scenarios. More detailed results are provided in Appendix~\ref{sec:appendixC.1}.

\input{table/deterministic_vs_VI}

\begin{figure}[t]
    \centering
    \includegraphics[width=1\textwidth]{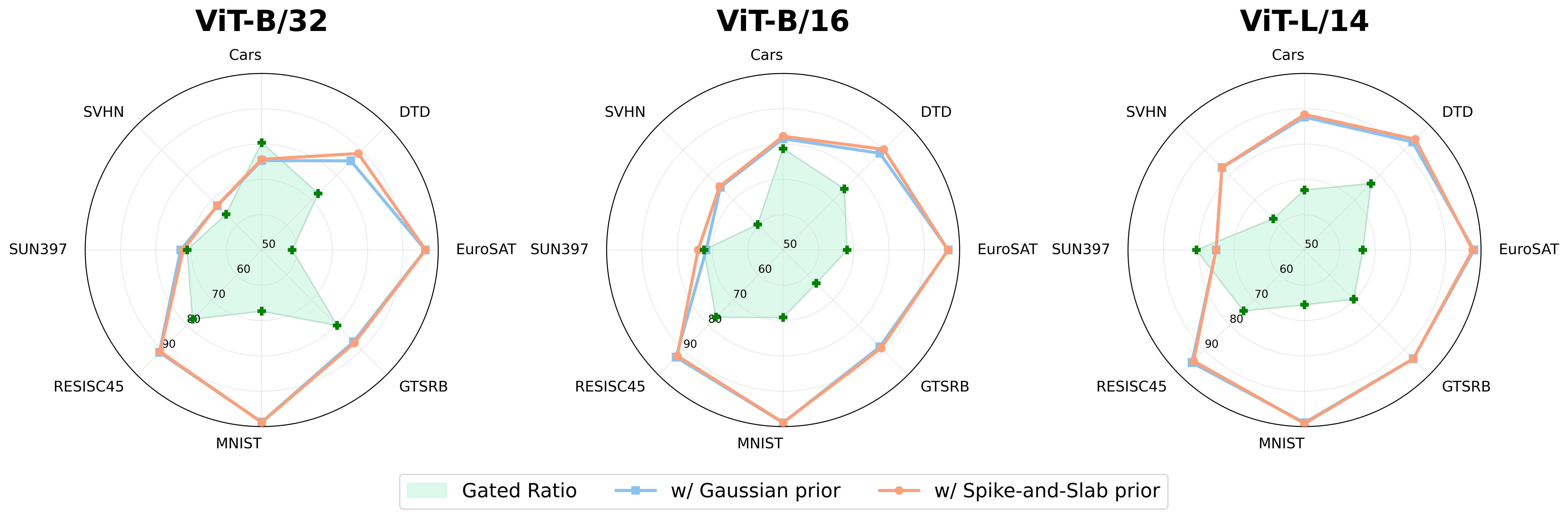}
    \caption{\textbf{Comparison of Accuracy and Gated Ratio under Gaussian and Spike-and-Slab Priors.} We visualize model accuracy and gated ratios for three ViT architectures across eight image classification datasets. The Spike-and-Slab prior consistently delivers higher accuracy while activating fewer task coefficients than the Gaussian prior, demonstrating more efficient and selective task vector composition.}
    \label{fig:N_vs_SpikeSlab}
    \vspace{-6mm}
\end{figure}

\paragraph{Normal Gaussian vs. Spike-Slab priors.} 
Figure~\ref{fig:N_vs_SpikeSlab} compares the performance of Gaussian and Spike-and-Slab priors across different ViT models. Spike-and-Slab priors achieve substantial performance gains on most datasets. These improvements are most evident in the datasets $\text{DTD}$, $\text{MNIST}$, and $\text{SVHN}$, which feature complex patterns or significant cross-domain differences. Spike-and-Slab priors excel at capturing sparse structures in combination with task vectors. They guide the model to select only the most relevant dimensions for each task. The gated ratio clearly demonstrates this effect. The increased sparsity reduces interference from redundant information and improves both efficiency and performance. It also reduces computational costs during inference. On $\text{EuroSAT}$ and $\text{RESISC45}$, Gaussian priors perform slightly better, possibly because these tasks benefit from dense, high-dimensional feature representations. Their tasks likely require comprehensive feature combinations across dimensions. Spike-and-Slab priors perform exceptionally well on datasets with complex and heterogeneous characteristics. The sparse approach improves cross-domain performance and makes the composition of task vectors more efficient. More detailed results are provided in Appendix~\ref{sec:appendixC.2}.

\paragraph{Effect of controllable posterior with gated sampling.} Table~\ref{tab:gated_vs_spikeslab} evaluates the effectiveness of the gated sampling mechanism. Introducing the gated sampling consistently improves performance over the baseline Spike-Slab prior. This improvement is particularly significant on datasets such as DTD, EuroSAT, and RESISC45. The gated sampling effectively identifies and preserves critical task knowledge while suppressing noise and unreliable components by evaluating the reliability of composition coefficients. The controllable posterior with gated sampling yields a more stable and interpretable component selection compared to inference with random sampling. Analysis in different data sets reveals that gated sampling performs more prominently on datasets with complex and variable feature spaces, further confirming its superiority in handling heterogeneous data. These experimental results validate the effectiveness of our theoretical framework, demonstrating the important role of uncertainty-guided adaptive component selection in enhancing model generalization capability and robustness. All results are based on features encoded by the $\text{ViT-B/32}$ model. More detailed results are provided in Appendix~\ref{sec:appendixC.3}. To further verify the effectiveness and transferability of the controllable posterior, we apply the gating module as a filter on the aTLAS method. The evaluation shows performance improvements across all datasets and three ViT architectures. Results are provided in Appendix~\ref{sec:appendixC.4}.

\input{table/gated_vs_spikeslab}

\begin{wrapfigure}{r}{0.48\textwidth}
\centering
\vspace{-7pt}
\includegraphics[width=0.48\textwidth]{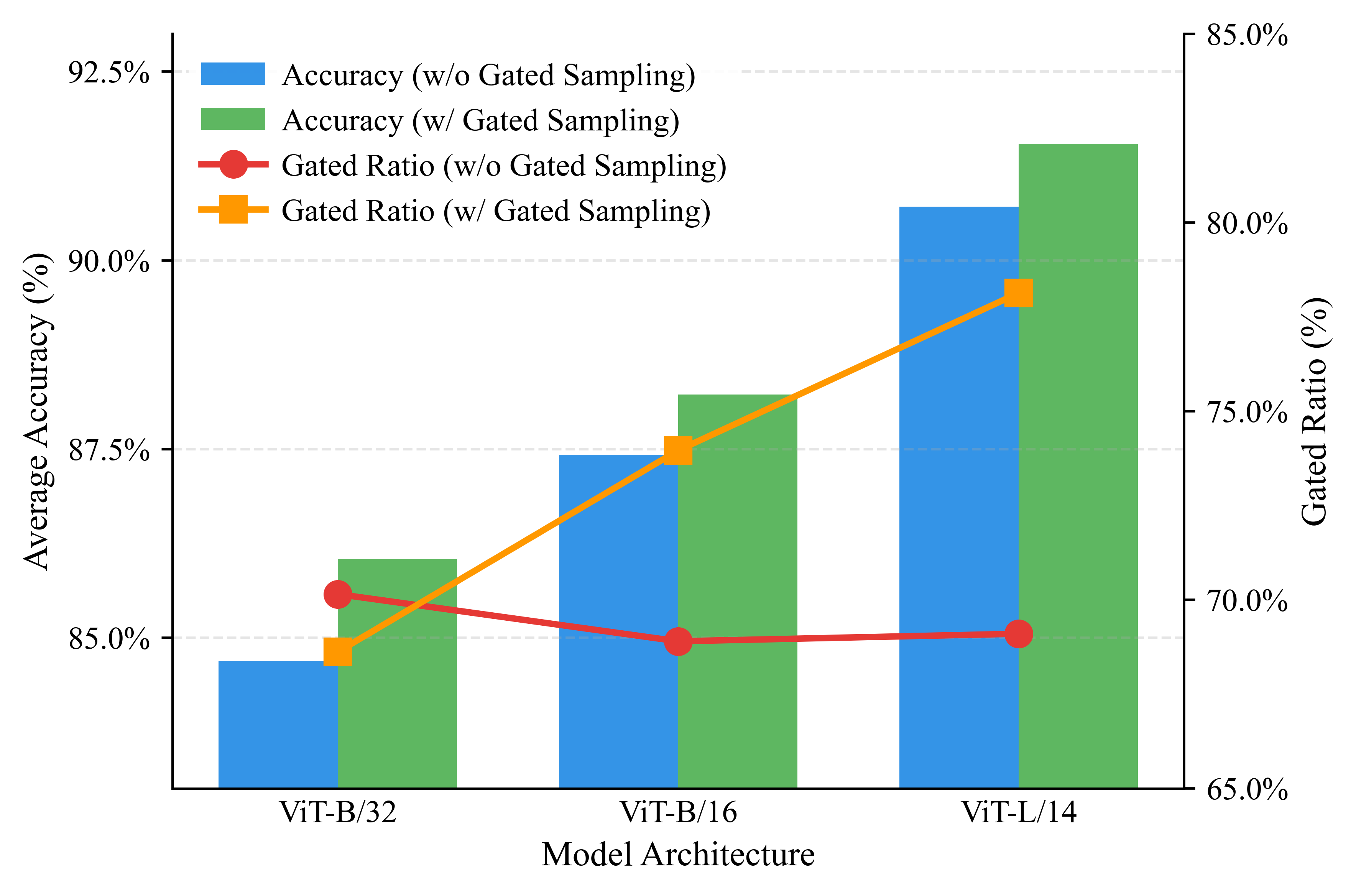}
\caption{\textbf{Comparison of Accuracy and Gated Ratio Across Model Architectures.} The gated sampling consistently improves prediction accuracy across all models. As model complexity increases, our framework retains more coefficients, showing the effectiveness of gated sampling in handling more complex tasks.}
\label{fig:gated_results}
\vspace{-8pt}
\end{wrapfigure}

\paragraph{Relationship Between Model Scale and Gated Ratio.} Figure~\ref{fig:gated_results} illustrates the impact of gated sampling on average accuracy and activation ratios across three Vision Transformer (ViT) models 
of increasing scale: $\text{ViT-B/32}$, $\text{ViT-B/16}$, and $\text{ViT-L/14}$. The results clearly demonstrate significant 
accuracy improvements when incorporating gated sampling in all models. The improvements are most pronounced in larger models ($\text{ViT-L/14}$), highlighting the scalability advantage of our method.
Notably, as the model scale increases, the activation coefficient ratio with gated posterior rises significantly, while models without gated sampling maintain relatively stable or slightly decreasing activation ratios. This pattern suggests conventional variational methods struggle to accommodate the knowledge integration demands of increasingly complex models. In contrast, our gated sampling dynamically adjusts knowledge integration strategies based on the model scale. It preserves more valuable task vectors in larger models, enabling more efficient and precise model editing. This phenomenon is consistent with the increased representational capacity and flexibility of larger models, which benefit more from adaptive gating strategies.

%% file: table/deterministic_vs_VI.tex
\begin{table}[t]
\centering
\caption{\textbf{Performance comparison of task-level versus sample-specific methods with deterministic and probabilistic task vectors across eight datasets.} Results demonstrate that sample-specific methods consistently outperform task-level methods across all datasets, with the variational inference framework further enhancing model performance.}
\small
\begin{adjustbox}{max width=\textwidth}
\begin{tabular}{cccccccccc}
\toprule
Method & Cars & DTD & EuroSAT & GTSRB & MNIST & RESISC45 & SUN397 & SVHN & Average \\
\midrule
Pre-trained & 58.56\% & 42.66\% & 40.74\% & 26.45\% & 25.23\% & 54.60\% & 46.46\% & 9.25\% & 37.99\% \\
\rowcolor[gray]{0.95}
\multicolumn{10}{c}{\textit{\small Deterministic-based}} \\
Task-level~\cite{knowledge_composition_2024}    & 62.99\% & 73.29\% & 92.67\% & 81.11\% & 97.13\% & 88.41\% & 67.16\% & 58.01\% & 77.60\% \\
\textbf{Sample-specific}  & 74.75\% & \textbf{85.97\%} & 96.07\% & 85.90\% & 98.07\% & 90.81\% & 68.01\% & 67.17\% & 83.34\% \\
\rowcolor[gray]{0.95}
\multicolumn{10}{c}{\textit{\small Probabilistic-based}} \\
Task-level    & 73.06\% & 80.64\% & 96.19\% & 85.66\% & 98.65\% & 90.00\% & 67.95\% & 64.23\% & 82.05\% \\
\textbf{Sample-specific}  & \textbf{75.33\%} & 85.71\% & \textbf{96.33\%} & \textbf{86.76\%} & \textbf{98.72\%} & \textbf{90.98\%} & \textbf{72.97\%} & \textbf{67.78\%} & \textbf{84.32\%} \\
\bottomrule
\label{tab:deterministic_vs_VI}
\end{tabular}
\end{adjustbox}
\vspace{-4mm}
\end{table}

%% file: table/gated_vs_spikeslab.tex
\begin{table}[t]
\centering
\caption{\textbf{Analysis of the adaptive gated sampling.} The gated sampling significantly improves model generalization capabilities across all datasets by selectively filtering reliable composition coefficients. }
\small
\begin{adjustbox}{max width=\textwidth}
\begin{tabular}{cccccccccc}
\toprule
Method & Cars & DTD & EuroSAT & GTSRB & MNIST & RESISC45 & SUN397 & SVHN & Average \\
\midrule
w/o Gated Sampling  & 75.69\% & 88.66\% & 96.41\% & \textbf{87.17\%} & 98.81\% & 90.71\% & 72.25\% & 67.85\% & 84.69\% \\
\textbf{w/ Gated Sampling} & \textbf{77.78\%} & \textbf{93.31\%} & \textbf{97.04\%} & 87.16\% & \textbf{98.91\%} & \textbf{92.08\%} & \textbf{73.47\%} & \textbf{68.57\%} & \textbf{86.04\%} \\
\bottomrule
\label{tab:gated_vs_spikeslab}
\end{tabular}
\end{adjustbox}
\vspace{-6mm}
\end{table}

%% file: sections/6_conclusion.tex
\section{Conclusion}
\label{sec: conclusion}

In this paper, we propose a new framework for the variational composition of task vectors by formulating the composition process as a variational inference problem. Our Bayesian approach enables precise, sample-specific modeling and effectively captures domain-specific knowledge for each sample. By introducing a Spike-and-Slab prior, our method automatically selects and preserves the most important components in task vectors while removing redundancy. Our gated sampling mechanism further enhances reliability by deterministically selecting the most critical task vectors. This approach avoids the randomness of conventional sampling methods and retains the strengths of variational inference. Extensive experiments demonstrate that our approach significantly improves model prediction performance and reduces spatial redundancy in task vectors. Overall, our framework offers a new perspective on task arithmetic and broadens the application potential of task vectors for model editing.

\noindent\textbf{Limitations.}
Since task addition serves as a broad and effective means to validate the utility of task vector composition, our work primarily focuses on this operation and does not explore other forms of task arithmetic. In addition, the high GPU memory requirements for combining multiple task vectors limit our ability to conduct experiments on larger-scale model architectures. Despite these constraints, our method demonstrates strong empirical performance and solid theoretical foundations on current models. We expect that the proposed framework can be extended to larger architectures. Additionally, exploring the applicability of our framework to other architectures (e.g., convolutional networks or language models) remains an interesting direction. In future work, we will further investigate the scalability and generalizability of our approach on larger and more complex models.

%% file: sections/7_appendix.tex
\section{Mathematical Derivation of Variational Composition of Task Vectors}
\label{sec:appendixA}
\subsection{Derivation of the ELBO Objective Function}
\label{sec:appendixA.1}
In the variational composition of task vectors, our main objective is to infer the posterior distribution $p(\mathbf{z}|\mathbf{x}, \mathbf{y})$, where $\mathbf{z}$ denotes the composition coefficients of task vectors, $\mathbf{x}$ denotes the input data, and $\mathbf{y}$ represents the labels. According to Bayes’ theorem, this posterior is given by:
\begin{equation}
    p(\mathbf{z}|\mathbf{x}, \mathbf{y}) = \frac{p(\mathbf{y}|\mathbf{x}, \mathbf{z})p(\mathbf{z})}{p(\mathbf{y}|\mathbf{x})}
\end{equation}
where the marginal likelihood $p(\mathbf{y}|\mathbf{x})$ is defined as:
\begin{equation}
    p(\mathbf{y}|\mathbf{x}) = \int p(\mathbf{y}|\mathbf{x}, \mathbf{z})p(\mathbf{z}) d\mathbf{z}
\end{equation}
Since directly computing the above integral is typically intractable, we introduce a variational approximate posterior distribution $q(\mathbf{z}|\mathbf{x})$ to approximate the true posterior. We derive a lower bound on the log marginal likelihood:
\begin{equation}
    \log p(\mathbf{y}|\mathbf{x}) = \log \int p(\mathbf{y}|\mathbf{x}, \mathbf{z})p(\mathbf{z}) d\mathbf{z}
\end{equation}
Introducing the variational distribution $q(\mathbf{z}|\mathbf{x})$:
\begin{equation}
    \begin{aligned}
        \log p(\mathbf{y}|\mathbf{x}) &= \log \int p(\mathbf{y}|\mathbf{x}, \mathbf{z})p(\mathbf{z}) d\mathbf{z} \\
        &= \log \int \frac{p(\mathbf{y}|\mathbf{x}, \mathbf{z})p(\mathbf{z})}{q(\mathbf{z}|\mathbf{x})}q(\mathbf{z}|\mathbf{x}) d\mathbf{z}
    \end{aligned}
\end{equation}
Since the logarithm is a concave function, by Jensen's inequality:
\begin{equation}
    \begin{aligned}
        \log p(\mathbf{y}|\mathbf{x}) &= \log \mathbb{E}_{q(\mathbf{z}|\mathbf{x})}\left[\frac{p(\mathbf{y}|\mathbf{x}, \mathbf{z})p(\mathbf{z})}{q(\mathbf{z}|\mathbf{x})}\right] \\
        &\geq \mathbb{E}_{q(\mathbf{z}|\mathbf{x})}\left[\log\frac{p(\mathbf{y}|\mathbf{x}, \mathbf{z})p(\mathbf{z})}{q(\mathbf{z}|\mathbf{x})}\right] \\
        &= \mathbb{E}_{q(\mathbf{z}|\mathbf{x})}[\log p(\mathbf{y}|\mathbf{x}, \mathbf{z})] + \mathbb{E}_{q(\mathbf{z}|\mathbf{x})}\left[\log\frac{p(\mathbf{z})}{q(\mathbf{z}|\mathbf{x})}\right] \\
        &= \mathbb{E}_{q(\mathbf{z}|\mathbf{x})}[\log p(\mathbf{y}|\mathbf{x}, \mathbf{z})] - D_\textbf{KL}(q(\mathbf{z}|\mathbf{x})||p(\mathbf{z}))
    \end{aligned}
\end{equation}
This derives Equation~\ref{Variational ELBO} from the paper:
\begin{equation}
    \log p(\mathbf{y}|\mathbf{x}) \geq \mathbb{E}_{q(\mathbf{z}|\mathbf{x})}[\log p(\mathbf{y}|\mathbf{x}, \mathbf{z})] - D_\textbf{KL}(q(\mathbf{z}|\mathbf{x})||p(\mathbf{z}))
\end{equation}
This inequality's right-hand side is known as the Evidence Lower BOund (ELBO). It comprises two critical components. The first term $\mathbb{E}_{q(\mathbf{z}|\mathbf{x})}[\log p(\mathbf{y}|\mathbf{x}, \mathbf{z})]$ represents the expected logarithmic likelihood, which guides the model toward accurate predictions. The second term $D_\textbf{KL}(q(\mathbf{z}|\mathbf{x})||p(\mathbf{z}))$ is the KL divergence term, regularizing the posterior toward the prior distribution, thus preventing overfitting.

Maximizing ELBO is equivalent to minimizing the KL divergence between the variational posterior and the true posterior. Through optimization of the ELBO, we can learn optimal composition coefficients for task vectors while maintaining robust generalization capabilities.

\subsection{Derivation of ELBO with Spike-and-Slab Prior}
\label{sec:appendixA.2}
In the Spike-and-Slab prior based variational framework, we introduce binary indicator variables $\Lambda$ to determine which composition coefficients are retained and which are set to zero. We define a joint variational posterior distribution $q_\Phi(\Lambda, \mathbf{z}|\mathbf{x})$, where $\Phi$ represents neural network parameters. The ELBO can be expressed as:
\begin{equation}
    \mathcal{L}_{\bf{ELBO}} = \mathbb{E}_{q_\Phi(\Lambda, \mathbf{z}|\mathbf{x})}[\log p(\mathbf{y}|\mathbf{x},\Lambda, \mathbf{z})] - D_{\textbf{KL}}(q_\Phi(\Lambda, \mathbf{z}|\mathbf{x})||p(\Lambda, \mathbf{z}))
\end{equation}
Using conditional probability decomposition, we can factorize the joint variational posterior as:
\begin{equation}
    q_\Phi(\Lambda, \mathbf{z}|\mathbf{x}) = q_\Phi(\mathbf{z}|\mathbf{x})q_\Phi(\Lambda|\mathbf{z},\mathbf{x})
\end{equation}
Similarly, the prior distribution can be decomposed as:
\begin{equation}
    p(\Lambda, \mathbf{z}) = p(\mathbf{z})p(\Lambda|\mathbf{z})
\end{equation}
Based on the above factorization, the KL divergence term can be further expanded:
\begin{equation}
    \begin{aligned}
        D_{\textbf{KL}}(q_\Phi(\Lambda, \mathbf{z}|\mathbf{x})||p(\Lambda, \mathbf{z})) &= D_{\textbf{KL}}(q_\Phi(\mathbf{z}|\mathbf{x})q_\Phi(\Lambda|\mathbf{z},\mathbf{x})||p(\mathbf{z})p(\Lambda|\mathbf{z})) \\
        &= \int q_\Phi(\mathbf{z}|\mathbf{x}) \log \frac{q_\Phi(\mathbf{z}|\mathbf{x})}{p(\mathbf{z})} d\mathbf{z} \\
        &+ \int q_\Phi(\mathbf{z}|\mathbf{x}) \left[ \int q_\Phi(\Lambda|\mathbf{z},\mathbf{x}) \log \frac{q_\Phi(\Lambda|\mathbf{z},\mathbf{x})}{p(\Lambda|\mathbf{z})} d\Lambda \right] d\mathbf{z} \\
        &= D_{\textbf{KL}}(q_\Phi(\mathbf{z}|\mathbf{x})||p(\mathbf{z})) + \mathbb{E}_{q_\Phi(\mathbf{z}|\mathbf{x})}[D_{\textbf{KL}}(q_\Phi(\Lambda|\mathbf{z},\mathbf{x})||p(\Lambda|\mathbf{z}))]
    \end{aligned}
\end{equation}

Therefore, the ELBO can be rewritten as:
\begin{equation}
    \mathcal{L}_{\bf{ELBO}} = \mathbb{E}_{q_\Phi(\Lambda, \mathbf{z}|\mathbf{x})}[\log p(\mathbf{y}|\mathbf{x},\Lambda, \mathbf{z})] - D_{\textbf{KL}}(q_\Phi(\mathbf{z}|\mathbf{x})||p(\mathbf{z})) - \mathbb{E}_{q_\Phi(\mathbf{z}|\mathbf{x})}[D_{\textbf{KL}}(q_\Phi(\Lambda|\mathbf{z},\mathbf{x})||p(\Lambda|\mathbf{z}))]
\end{equation}
In our model, we formulate a comprehensive probabilistic framework by defining the following distribution structure: The binary variables in the variational posterior follow $q_\Phi(\Lambda_i^j|\mathbf{x}) = \text{Bernoulli}(\gamma_i^j(\mathbf{x}))$, where $\gamma_i^j(\mathbf{x})$ represents the inclusion probability for each coefficient, while their prior counterparts are distributed according to $p(\Lambda_i^j) = \text{Bernoulli}(\pi)$, with $\pi$ serving as the global sparsity parameter. For the coefficient values themselves, when $\Lambda_i^j = 1$, the posterior distribution is characterized by $q_\Phi(\mathbf{z}_i^j|\Lambda_i^j=1,\mathbf{x}) = \mathcal{N}(\mathbf{z}_i^j; \mu_{ij}(\mathbf{x}), \sigma_{ij}^2(\mathbf{x}))$, and correspondingly, the prior follows $p(\mathbf{z}_i^j|\Lambda_i^j=1) = \mathcal{N}(\mathbf{z}_i^j; 0, \sigma^2)$, establishing a complete probabilistic specification of our variable selection mechanism.

For the binary indicator variables $\Lambda$, the KL divergence is calculated as:
\begin{equation}
    \begin{aligned}
        D_{\textbf{KL}}(q_\Phi(\Lambda|\mathbf{x})||p(\Lambda)) &= \sum_{ij} D_{\textbf{KL}}(\text{Bernoulli}(\gamma_i^j(\mathbf{x}))||\text{Bernoulli}(\pi)) \\
        &= \sum_{ij} \gamma_i^j(\mathbf{x}) \log\frac{\gamma_i^j(\mathbf{x})}{\pi} + (1-\gamma_i^j(\mathbf{x})) \log\frac{1-\gamma_i^j(\mathbf{x})}{1-\pi}
    \end{aligned}
\end{equation}
When $\Lambda_i^j = 1$, the KL divergence for coefficient $z_i^j$ is:
\begin{equation}
    \begin{aligned}
        D_{\textbf{KL}}(q_\Phi(\mathbf{z}_i^j|\Lambda_i^j=1,\mathbf{x})||p(\mathbf{z}_i^j|\Lambda_i^j=1)) &= D_{\textbf{KL}}(\mathcal{N}(\mathbf{z}_i^j; \mu_{ij}(\mathbf{x}), \sigma_{ij}^2(\mathbf{x}))||\mathcal{N}(\mathbf{z}_i^j; 0, \sigma^2)) \\
        &= \frac{1}{2} \left[ \log\frac{\sigma^2}{\sigma_{ij}^2(\mathbf{x})} + \frac{\sigma_{ij}^2(\mathbf{x}) + \mu_{ij}^2(\mathbf{x})}{\sigma^2} - 1 \right]
    \end{aligned}
\end{equation}
Combining the above KL divergence calculations, and considering that when $\Lambda_i^j = 0$, the coefficient is forced to zero (thus contributing no additional KL divergence), the complete form of the ELBO is:
\begin{equation}
    \begin{aligned}
        \mathcal{L}_{\bf{ELBO}} &= \mathbb{E}_{q_\Phi(\Lambda, \mathbf{z}|\mathbf{x})}[\log p(\mathbf{y}|\mathbf{x},\Lambda, \mathbf{z})] \\
        &- \left[ \sum_{ij}\gamma_i^j(\mathbf{x})\log\frac{\gamma_i^j(\mathbf{x})}{\pi} + \sum_{ij}(1-\gamma_i^j(\mathbf{x}))\log\frac{1-\gamma_i^j(\mathbf{x})}{1-\pi} \right] \\
        &- \sum_{ij}\gamma_i^j(\mathbf{x}) \cdot \frac{1}{2} \left[ \log\frac{\sigma^2}{\sigma_{ij}^2(\mathbf{x})} + \frac{\sigma_{ij}^2(\mathbf{x}) + \mu_{ij}^2(\mathbf{x})}{\sigma^2} - 1 \right]
    \end{aligned}
\end{equation}
This represents the complete expansion of Equation~\ref{variational posterior}. The first term is the expected log-likelihood, representing the model's predictive accuracy; the second term is the KL divergence of binary indicator variables, controlling the sparsity pattern; and the third term is the conditional KL divergence of non-zero coefficients, weighted by their inclusion probabilities, ensuring a reasonable distribution of weight values. By optimizing this objective function, we achieve adaptive sparse composition of task vectors while maintaining the model's generalization capability.

\section{Regularization and Hyperparameter Settings}
\label{sec:appendixB}
\noindent\textbf{Regularization Overview}. In our adaptive gating mechanism, regularization plays a crucial role in ensuring training stability and enhancing model generalization capability. Our proposed regularization strategy addresses several key challenges: uncertainty in gating decisions, insufficient parameter space exploration, and information integration efficiency. When coefficient values approach thresholds, small perturbations can cause unstable selection results; training processes tend to converge prematurely to suboptimal solutions; and selected components must effectively represent sample features. The total regularization objective $\mathcal{L}_{reg}(\mathbf{z}, \mathcal{U}; \Psi)$ comprises multiple carefully designed components, each aiming to jointly improve the robustness and effectiveness of the gating mechanism.

\noindent\textbf{Boundary Loss}. The boundary loss $\mathcal{L}_{b}$ is specifically designed to penalize coefficients near thresholds, enhancing stability by encouraging more decisive selection decisions. When coefficient values lie in ambiguous threshold regions, minor input variations can cause significant fluctuations in component selection, reducing model robustness and generalization capacity. The boundary loss is defined as:
\begin{equation}
    \mathcal{L}_{b}(\mathbf{z}, \Psi) = \sum_{i=1}^{n} \sum_{j=1}^{b} \max(0, m - |\mathbf{z}_{i,j} - \Gamma(U_{i,j}; \Psi)|)
\end{equation}
This loss function imposes a positive penalty when the absolute difference between coefficient $\mathbf{z}_{i,j}$ and threshold $\Gamma(U_{i,j}; \Psi)$ falls below boundary width $m$, otherwise zero, effectively pushing coefficients away from unstable threshold regions to form more decisive binary selection patterns. In our experiments, the default boundary width parameter $m=0.1$ effectively balances clear decision-making with model flexibility.

\noindent\textbf{Exploration Loss}. Exploration loss $\mathcal{L}_{e}$ promotes threshold-related parameters ($\psi_1$ and $\psi_2$) to explore broader parameter spaces, preventing training from converging to local optima. By encouraging these parameters to deviate from their initial values, the model explores potentially superior parameter configurations, enhancing the adaptability of the gating mechanism. The exploration loss is defined as:
\begin{equation}
    \mathcal{L}_{e}(\Psi) = -\log\left(|\psi_1 - \psi_1^{0}| + \epsilon\right) - \log\left(|\psi_2 - \psi_2^{0}| + \epsilon\right)
\end{equation}
This loss function decreases as parameters deviate from their initial configurations by taking the negative logarithm of the parameter-initial value difference, thus incentivizing broader parameter space exploration. Here, $\epsilon$ represents a small constant (typically $10^{-5}$) that ensures numerical stability. The uncertainty loss $\mathcal{L}_{u}$ is defined as:
\begin{equation}
    \mathcal{L}_{u}(\mathcal{U}) = \sum_{i=1}^{n} \sum_{j=1}^{b} \mathcal{U}_{i,j} \cdot \mathbb{I}[\mathbf{z}_{i,j} \neq 0]
\end{equation}
This loss penalizes selected (non-zero) coefficients with high uncertainty, encouraging the model to reduce uncertainty while retaining coefficients, further enhancing gating decision reliability. The comprehensive regularization objective integrates these components:
\begin{equation}
    \mathcal{L}_{reg}(\mathbf{z}, \mathcal{U}; \Psi) = \lambda_{b} \mathcal{L}_{b}(\mathbf{z}, \Psi) + \lambda_{e} \mathcal{L}_{e}(\Psi) + \lambda_{u} \mathcal{L}_{u}(\mathcal{U})
\end{equation}
\noindent\textbf{Hyperparameter Settings and Sensitivity Analysis}. In our experimental implementation, we used a set of meticulously tuned hyperparameters: boundary width $m = 0.1$, boundary loss weight $\lambda_{b} = 10^{-4}$, exploration loss weight $\lambda_{e} = 10^{-3}$, uncertainty loss weight $\lambda_{u} = 10^{-2}$, base threshold initial value $\psi_1^{0} = 0.05$, uncertainty sensitivity initial value $\psi_2^{0} = 1.0$, global regularization coefficient $\lambda = 10^{-3}$ (Equation~\ref{gated posterior}), and gating temperature parameter $\rho = 20.0$ (Equation~\ref{selection mechanism}). Sensitivity analysis indicates that model performance is relatively robust to variations in $\lambda_{b}$ within the range $[10^{-5}, 10^{-3}]$; more sensitive to $\lambda_{e}$, where increasing this value ($>10^{-2}$) promotes more aggressive parameter exploration but can lead to training instability, while decreasing it ($<10^{-4}$) can cause the model to converge to local optima. Performance is optimal when $\lambda_{u}$ is within $[5\times10^{-3}, 5\times10^{-2}]$. The initial setting of the base threshold $\psi_1^{0}$ significantly impacts model performance: excessive values ($>0.1$) lead to overly sparse representations, compromising information retention; insufficient values ($<0.01$) retain too many redundant components, increasing overfit risk. For different feature dimensions and dataset characteristics, we recommend setting $\psi_1^{0}$ within the range $[0.03, 0.08]$ and automatically selecting optimal configurations through validation sets. Comprehensive experimental results demonstrate that our proposed regularization strategy and parameter configurations significantly enhance the effectiveness of the adaptive gating mechanism, enabling reliable component selection across diverse task vector spaces.

\section{Supplementary Experimental Results}
\label{sec:appendixC}

\subsection{Detailed results of task-level vs. sample-specific and deterministic vs. probabilistic}
\label{sec:appendixC.1}
\input{table/full_comparison_results}

\subsection{Detailed results of Gaussian vs. Spike-and-Slab priors}
\label{sec:appendixC.2}
\input{table/N_spike_slab_comparison}

\subsection{Detailed results of gated sampling}
\label{sec:appendixC.3}
\input{table/gated_ratio_results}

\subsection{Applying gated sampling to aTLAS for filtering}
\label{sec:appendixC.4}
\input{table/aTLAS_with_gating}

%% file: table/full_comparison_results.tex
\begin{table}[htb]
\centering
\caption{\textbf{Detailed accuracy comparison (\%) across ViT models and datasets.} We report zero-shot, deterministic (task-level and sample-specific), and probabilistic (task-level and sample-specific) performances for each model. Highest performance in each dataset is highlighted in \textbf{bold}.}
\footnotesize
\begin{adjustbox}{max width=\textwidth}
\begin{tabular}{lcccccccccc}
\toprule
Method & Cars & DTD & EuroSAT & GTSRB & MNIST & RESISC45 & SUN397 & SVHN & Avg. \\
\midrule
\rowcolor[gray]{0.85}
\multicolumn{10}{c}{\textit{\textbf{ViT-B/32}}} \\
Pre-trained & 58.56 & 42.66 & 40.74 & 26.45 & 25.23 & 54.60 & 46.46 & 9.25 & 37.99 \\
\rowcolor[gray]{0.95}
\multicolumn{10}{c}{\textit{Deterministic-based}} \\
Task-level & 62.99 & 73.29 & 92.67 & 81.11 & 97.13 & 88.41 & 67.16 & 58.01 & 77.60 \\
Sample-specific & 74.75 & \textbf{85.97} & 96.07 & 85.90 & 98.07 & 90.81 & 68.01 & 67.17 & 83.34 \\
\rowcolor[gray]{0.95}
\multicolumn{10}{c}{\textit{Probabilistic-based}} \\
Task-level & 73.06 & 80.64 & 96.19 & 85.66 & 98.65 & 90.00 & 67.95 & 64.23 & 82.05 \\
Sample-specific & \textbf{75.33} & 85.71 & \textbf{96.33} & \textbf{86.76} & \textbf{98.72} & \textbf{90.98} & \textbf{72.97} & \textbf{67.78} & \textbf{84.32} \\
\midrule
\rowcolor[gray]{0.85}
\multicolumn{10}{c}{\textit{\textbf{ViT-B/16}}} \\
Zero-shot & 63.45 & 43.77 & 50.63 & 35.22 & 19.29 & 55.70 & 47.13 & 15.53 & 41.34 \\
\rowcolor[gray]{0.95}
\multicolumn{10}{c}{\textit{Deterministic-based}} \\
Task-level & 71.61 & 75.92 & 93.56 & 82.15 & 97.36 & 90.37 & 69.54 & 66.67 & 80.90 \\
Sample-specific & 80.59 & \textbf{90.10} & 96.96 & 88.04 & 98.85 & 92.17 & 70.02 & 74.97 & 86.46 \\
\rowcolor[gray]{0.95}
\multicolumn{10}{c}{\textit{Probabilistic-based}} \\
Task-level & 81.32 & 82.44 & 96.19 & 88.18 & 98.93 & 91.57 & 69.95 & 70.50 & 84.87 \\
Sample-specific & \textbf{81.52} & 88.75 & \textbf{96.78} & \textbf{88.76} & \textbf{98.93} & \textbf{92.90} & \textbf{71.81} & \textbf{75.12} & \textbf{86.82} \\
\midrule
\rowcolor[gray]{0.85}
\multicolumn{10}{c}{\textit{\textbf{ViT-L/14}}} \\
Zero-shot & 76.40 & 52.10 & 55.63 & 45.61 & 52.59 & 63.41 & 50.41 & 41.73 & 54.74 \\
\rowcolor[gray]{0.95}
\multicolumn{10}{c}{\textit{Deterministic-based}} \\
Task-level & 85.39 & 83.41 & 96.89 & 91.55 & 98.76 & 93.79 & 71.05 & 76.02 & 87.11 \\
Sample-specific & 87.29 & \textbf{96.84} & 97.91 & 93.46 & \textbf{99.09} & 94.30 & 73.90 & 82.38 & 90.65 \\
\rowcolor[gray]{0.95}
\multicolumn{10}{c}{\textit{Probabilistic-based}} \\
Task-level & 88.47 & 90.08 & 97.78 & 92.87 & 99.01 & 93.54 & 72.33 & 78.46 & 89.07 \\
Sample-specific & \textbf{89.64} & 93.28 & \textbf{97.99} & \textbf{93.55} & 98.92 & \textbf{95.11} & \textbf{75.05} & \textbf{83.03} & \textbf{90.82} \\
\bottomrule
\end{tabular}
\end{adjustbox}
\label{tab:full_comparison_results}
\end{table}

%% file: table/N_spike_slab_comparison.tex
\begin{table}[htb]
\centering
\caption{\textbf{Performance comparison between Gaussian and Spike-and-Slab prior models (\%) across ViT models and datasets.} Highest performance in each dataset is highlighted in \textbf{bold}. Gated ratio values are reported in \textit{italics}.}
\footnotesize
\begin{adjustbox}{max width=\textwidth}
\begin{tabular}{lcccccccccc}
\toprule
Method & Cars & DTD & EuroSAT & GTSRB & MNIST & RESISC45 & SUN397 & SVHN & Avg. \\
\midrule
\rowcolor[gray]{0.85}
\multicolumn{10}{c}{\textit{\textbf{ViT-B/32}}} \\
Gaussian prior & 75.33\% & 85.71\% & 96.33\% & 86.76\% & 98.72\% & \textbf{90.98\%} & \textbf{72.97\%} & 67.78\% & 84.32\% \\
Spike-and-Slab prior & \textbf{75.69\%} & \textbf{88.66\%} & \textbf{96.41\%} & \textbf{87.17\%} & \textbf{98.81\%} & 90.71\% & 72.25\% & \textbf{67.85\%} & 84.69\% \\
\midrule
\rowcolor[gray]{0.85}
\multicolumn{10}{c}{\textit{\textbf{ViT-B/16}}} \\
Gaussian prior & 81.52\% & 88.75\% & \textbf{96.78\%} & 88.76\% & 98.85\% & 92.17\% & 71.81\% & 74.97\% & 86.61\% \\
Spike-and-Slab prior & \textbf{82.23\%} & \textbf{90.31\%} & 96.70\% & \textbf{89.27\%} & \textbf{98.87\%} & \textbf{92.40\%} & \textbf{74.10\%} & \textbf{75.46\%} & \textbf{87.42\%} \\
\rowcolor[gray]{0.95}
\midrule
\rowcolor[gray]{0.85}
\multicolumn{10}{c}{\textit{\textbf{ViT-L/14}}} \\
Gaussian prior & 87.29\% & 93.28\% & \textbf{97.67\%} & 93.46\% & 99.09\% & 94.30\% & 73.90\% & 82.38\% & 90.38\% \\
Spike-and-Slab prior & \textbf{88.38\%} & \textbf{94.35\%} & 97.63\% & \textbf{93.56\%} & \textbf{99.14\%} & \textbf{94.43\%} & \textbf{75.07\%} & \textbf{83.09\%} & \textbf{90.71\%} \\
\rowcolor[gray]{0.95}
\bottomrule
\end{tabular}
\end{adjustbox}
\label{tab:N_spike_slab_comparison}
\end{table}

%% file: table/gated_ratio_results.tex
\begin{table}[H]
\centering
\caption{\textbf{Accuracy and gating ratio comparison (\%) across ViT models and datasets.} Performance and average gated ratio are reported for models with and without gated sampling. The highest performance in each dataset is highlighted in \textbf{bold}. Gated ratio values are in \textit{italics}.}
\scalebox{0.8}{
\begin{tabularx}{1.25\textwidth}{l *{9}{>{\centering\arraybackslash}X}}
\toprule
Method & Cars & DTD & EuroSAT & GTSRB & MNIST & RESISC45 &  SUN397 & SVHN & Avg. \\
\midrule
\rowcolor[gray]{0.85}
\multicolumn{10}{c}{\textit{\textbf{ViT-B/32}}} \\
w/o Gated Sampling & 75.69\% & 88.66\% & 96.41\% & \textbf{87.17\%} & 98.81\% & 90.71\% & 72.25\% & 67.85\% & 84.69\% \\
\rowcolor[gray]{0.95}
Gated Ratio & \underline{\textit{80.40\%}} & \underline{\textit{72.60\%}} & \underline{\textit{58.60\%}} & \underline{\textit{80.20\%}} & \underline{\textit{67.30\%}} & \underline{\textit{77.70\%}} & \underline{\textit{71.10\%}} & \underline{\textit{64.30\%}} & \underline{\textit{71.53\%}} \\
w/ Gated Sampling & \textbf{77.78\%} & \textbf{93.31\%} & \textbf{97.04\%} & 87.16\% & \textbf{98.91\%} & \textbf{92.08\%} & \textbf{73.47\%} & \textbf{68.57\%} & \textbf{86.04\%} \\
\rowcolor[gray]{0.95}
Gated Ratio & \underline{\textit{65.62\%}} & \underline{\textit{71.88\%}} & \underline{\textit{59.38\%}} & \underline{\textit{71.87\%}} & \underline{\textit{60.42\%}} & \underline{\textit{63.54\%}} & \underline{\textit{91.67\%}} & \underline{\textit{64.58\%}} & \underline{\textit{68.62\%}} \\
\midrule
\rowcolor[gray]{0.85}
\multicolumn{10}{c}{\textit{\textbf{ViT-B/16}}} \\
w/o Gated Sampling & 82.23\% & 90.31\% & 96.70\% & \textbf{89.27\%} & 98.87\% & 92.40\% & \textbf{74.10\%} & 75.46\% & 87.42\% \\
\rowcolor[gray]{0.95}
Gated Ratio & \underline{\textit{78.70\%}} & \underline{\textit{74.50\%}} & \underline{\textit{68.10\%}} & \underline{\textit{63.30\%}} & \underline{\textit{69.10\%}} & \underline{\textit{76.90\%}} & \underline{\textit{72.40\%}} & \underline{\textit{60.20\%}} & \underline{\textit{70.40\%}} \\
w/ Gated Sampling & \textbf{83.61\%} & \textbf{95.01\%} & \textbf{97.00\%} & 89.22\% & \textbf{98.93\%} & \textbf{93.02\%} & 73.53\% & \textbf{75.47\%} & \textbf{88.22\%} \\
\rowcolor[gray]{0.95}
Gated Ratio & \underline{\textit{79.17\%}} & \underline{\textit{73.96\%}} & \underline{\textit{67.71\%}} & \underline{\textit{82.23\%}} & \underline{\textit{81.25\%}} & \underline{\textit{77.08\%}} & \underline{\textit{62.08\%}} & \underline{\textit{75.21\%}} & \underline{\textit{75.21\%}} \\
\midrule
\rowcolor[gray]{0.85}
\multicolumn{10}{c}{\textit{\textbf{ViT-L/14}}} \\
w/o Gated Sampling & 88.38\% & 94.35\% & 97.63\% & 93.56\% & 99.14\% & 94.43\% & 75.07\% & 83.09\% & 90.71\% \\
\rowcolor[gray]{0.95}
Gated Ratio & \underline{\textit{67.00\%}} & \underline{\textit{76.60\%}} & \underline{\textit{66.50\%}} & \underline{\textit{69.70\%}} & \underline{\textit{65.50\%}} & \underline{\textit{74.40\%}} & \underline{\textit{62.50\%}} & \underline{\textit{80.60\%}} & \underline{\textit{70.35\%}} \\
w/ Gated Sampling & \textbf{89.45\%} & \textbf{97.75\%} & \textbf{97.96\%} & \textbf{93.67\%} & \textbf{99.15\%} & \textbf{95.49\%} & \textbf{75.52\%} & \textbf{83.31\%} & \textbf{91.54\%} \\
\rowcolor[gray]{0.95}
Gated Ratio & \underline{\textit{69.79\%}} & \underline{\textit{72.92\%}} & \underline{\textit{70.83\%}} & \underline{\textit{85.83\%}} & \underline{\textit{85.42\%}} & \underline{\textit{85.42\%}} & \underline{\textit{67.71\%}} & \underline{\textit{78.13\%}} & \underline{\textit{78.13\%}} \\
\bottomrule
\end{tabularx}}
\label{tab:gated_ratio_results}
\end{table}

%% file: table/aTLAS_with_gating.tex
\begin{table}[htb]
\centering
\caption{\textbf{Performance comparison between standard aTLAS and Gated aTLAS methods across three ViT architectures and eight datasets (\%).} The gated mechanism consistently improves performance across all datasets and models. Highest performance in each dataset is highlighted in \textbf{bold}.}
\small
\begin{adjustbox}{max width=\textwidth}
\begin{tabular}{lcccccccccc}
\toprule
Method & Cars & DTD & EuroSAT & GTSRB & MNIST & RESISC45 & SUN397 & SVHN & Avg. \\
\midrule
\rowcolor[gray]{0.85}
\multicolumn{10}{c}{\textbf{\textit{ViT-B/32}}} \\
w/o Gated & 62.99\% & 73.29\% & 92.67\% & 81.11\% & 97.13\% & 88.41\% & 67.16\% & 58.01\% & 77.60\% \\
w/ Gated & \textbf{71.32\%} & \textbf{80.24\%} & \textbf{94.96\%} & \textbf{84.92\%} & \textbf{97.96\%} & \textbf{89.84\%} & \textbf{68.33\%} & \textbf{61.68\%} & \textbf{81.16\%} \\
\rowcolor[gray]{0.85}
\multicolumn{10}{c}{\textbf{\textit{ViT-B/16}}} \\
w/o Gated & 71.61\% & 75.92\% & 93.56\% & 82.15\% & 97.36\% & 90.37\% & 69.54\% & 66.67\% & 80.90\% \\
w/ Gated & \textbf{80.06\%} & \textbf{82.37\%} & \textbf{94.96\%} & \textbf{86.04\%} & \textbf{98.10\%} & \textbf{91.59\%} & \textbf{70.29\%} & \textbf{68.68\%} & \textbf{84.01\%} \\
\rowcolor[gray]{0.85}
\multicolumn{10}{c}{\textbf{\textit{ViT-L/14}}} \\
w/o Gated & 85.39\% & 83.41\% & 96.89\% & 91.55\% & 98.76\% & 93.79\% & 73.05\% & 76.02\% & 87.11\% \\
w/ Gated & \textbf{86.78\%} & \textbf{88.40\%} & \textbf{97.33\%} & \textbf{92.88\%} & \textbf{98.90\%} & \textbf{94.75\%} & \textbf{77.19\%} & \textbf{77.79\%} & \textbf{88.59\%} \\
\bottomrule
\end{tabular}
\end{adjustbox}
\label{tab:aTLAS_with_gating}
\end{table}

%% file: sections/8_checklist.tex
\newpage
\section*{NeurIPS Paper Checklist}

\begin{enumerate}

\item {\bf Claims}
    \item[] Question: Do the main claims made in the abstract and introduction accurately reflect the paper's contributions and scope?
    \item[] Answer: \answerYes{} 
    \item[] Justification: 
    The contributions and scope of this paper are claimed in the abstract. Detailed information can be found in the third paragraph of the introduction section~\ref{sec: introduction}.

\item {\bf Limitations}
    \item[] Question: Does the paper discuss the limitations of the work performed by the authors?
    \item[] Answer: \answerYes{}
    \item[] Justification: 
    We provide a "limitation" subsection in the conclusion section~\ref{sec: conclusion}.

\item {\bf Theory Assumptions and Proofs}
    \item[] Question: For each theoretical result, does the paper provide the full set of assumptions and a complete (and correct) proof?
    \item[] Answer: \answerNA{}
    \item[] Justification: 
    This paper does not include theoretical results.

    \item {\bf Experimental Result Reproducibility}
    \item[] Question: Does the paper fully disclose all the information needed to reproduce the main experimental results of the paper to the extent that it affects the main claims and/or conclusions of the paper (regardless of whether the code and data are provided or not)?
    \item[] Answer: \answerYes{}
    \item[] Justification: 
    We provide all experimental details in the experiment section. The detailed input and output of our method can be found in Appendix and supplemental materials.

\item {\bf Open access to data and code}
    \item[] Question: Does the paper provide open access to the data and code, with sufficient instructions to faithfully reproduce the main experimental results, as described in supplemental material?
    \item[] Answer: 
    \answerYes{}
    \item[] Justification: 
    We provide all experimental details in the experiment section. The detailed and Python codes of our method can be found in Appendix and supplemental materials.

\item {\bf Experimental Setting/Details}
    \item[] Question: Does the paper specify all the training and test details (e.g., data splits, hyperparameters, how they were chosen, type of optimizer, etc.) necessary to understand the results?
    \item[] Answer: 
    \answerYes{}
    \item[] Justification: 
    We follow the standard experimental setup in \ours, where the data splits, is set as the same as previous works aTLAS~\cite{knowledge_composition_2024}. Detailed information can be found in Sec.~\ref{sec: experiment}

\item {\bf Experiment Statistical Significance}
    \item[] Question: Does the paper report error bars suitably and correctly defined or other appropriate information about the statistical significance of the experiments?
    \item[] Answer: \answerYes{}
    \item[] Justification: 
     Following the standard experimental setup, we repeat each experiment over 3 random seeds and report the mean of the results.

\item {\bf Experiments Compute Resources}
    \item[] Question: For each experiment, does the paper provide sufficient information on the computer resources (type of compute workers, memory, time of execution) needed to reproduce the experiments?
    \item[] Answer: \answerYes{}
    \item[] Justification: 
    We provide the computing resources in experiments~\ref{sec: experiment}.
    
\item {\bf Code Of Ethics}
    \item[] Question: Does the research conducted in the paper conform, in every respect, with the NeurIPS Code of Ethics \url{https://neurips.cc/public/EthicsGuidelines}?
    \item[] Answer: \answerYes{}
    \item[] Justification: 
    We reviewed and followed the NeurIPS Code of Ethics.

\item {\bf Broader Impacts}
    \item[] Question: Does the paper discuss both potential positive societal impacts and negative societal impacts of the work performed?
    \item[] Answer: \answerYes{}
    \item[] Justification: 
    We provide the potential broader impacts in the conclusion section~\ref{sec: conclusion}.
    
\item {\bf Safeguards}
    \item[] Question: Does the paper describe safeguards that have been put in place for responsible release of data or models that have a high risk for misuse (e.g., pretrained language models, image generators, or scraped datasets)?
    \item[] Answer: \answerNA{}
    \item[] Justification: The data and models pose no such risks.

\item {\bf Licenses for existing assets}
    \item[] Question: Are the creators or original owners of assets (e.g., code, data, models), used in the paper, properly credited and are the license and terms of use explicitly mentioned and properly respected?
    \item[] Answer: \answerYes{}
    \item[] Justification:
    We cite the original papers that produced the code package and datasets.

\item {\bf New Assets}
    \item[] Question: Are new assets introduced in the paper well documented and is the documentation provided alongside the assets?
    \item[] Answer: \answerYes{}
    \item[] Justification: 
    Details of the datasets/code/model are provided in the supplemental materials. 

\item {\bf Crowdsourcing and Research with Human Subjects}
    \item[] Question: For crowdsourcing experiments and research with human subjects, does the paper include the full text of instructions given to participants and screenshots, if applicable, as well as details about compensation (if any)? 
    \item[] Answer: \answerNA{}
    \item[] Justification: 
    This paper does not involve crowdsourcing nor research with human subjects.

\item {\bf Institutional Review Board (IRB) Approvals or Equivalent for Research with Human Subjects}
    \item[] Question: Does the paper describe potential risks incurred by study participants, whether such risks were disclosed to the subjects, and whether Institutional Review Board (IRB) approvals (or an equivalent approval/review based on the requirements of your country or institution) were obtained?
    \item[] Answer: \answerNA{}
    \item[] Justification: 
    This paper does not involve crowdsourcing nor research with human subjects.

\item {\bf Declaration of LLM usage}
    \item[] Question: Does the paper describe the usage of LLMs if it is an important, original, or non-standard component of the core methods in this research? Note that if the LLM is used only for writing, editing, or formatting purposes and does not impact the core methodology, scientific rigorousness, or originality of the research, declaration is not required.
    \item[] Answer: \answerYes{} 
    \item[] Justification: We use LLM solely for grammar checking.

\end{enumerate}

%% file: main.bbl
\begin{thebibliography}{10}

\bibitem{bayesian_spike_slab_2014}
M.~R. Andersen, O.~Winther, and L.~K. Hansen.
\newblock Bayesian inference for structured spike and slab priors.
\newblock {\em Advances in Neural Information Processing Systems}, 27, 2014.

\bibitem{spike_slab_lasso_2021}
R.~Bai, V.~Ro{\v{c}}kov{\'a}, and E.~I. George.
\newblock Spike-and-slab meets lasso: A review of the spike-and-slab lasso.
\newblock {\em Handbook of Bayesian variable selection}, pages 81--108, 2021.

\bibitem{resisc45_2017}
G.~Cheng, J.~Han, and X.~Lu.
\newblock Remote sensing image scene classification: Benchmark and state of the art.
\newblock {\em Proceedings of IEEE}, 105(10):1865--1883, Oct 2017.

\bibitem{dtd_2014}
M.~Cimpoi, S.~Maji, I.~Kokkinos, S.~Mohamed, and A.~Vedaldi.
\newblock Describing textures in the wild.
\newblock In {\em Proceedings of IEEE/CVF Conference on Computer Vision and Pattern Recognition (CVPR)}, pages 3606--3613, Columbus, OH, USA, 24--27 Jun 2014.

\bibitem{fast_spike_slab_2022}
H.~Dance and B.~Paige.
\newblock Fast and scalable spike and slab variable selection in high-dimensional gaussian processes.
\newblock In {\em International Conference on Artificial Intelligence and Statistics}, pages 7976--8002. PMLR, 2022.

\bibitem{challenges_high_dim_2021}
A.~K. Dhaka, A.~Catalina, M.~Welandawe, M.~R. Andersen, J.~Huggins, and A.~Vehtari.
\newblock Challenges and opportunities in high dimensional variational inference.
\newblock {\em Advances in Neural Information Processing Systems}, 34:7787--7798, 2021.

\bibitem{vit_2021}
A.~Dosovitskiy, L.~Beyer, A.~Kolesnikov, D.~Weissenborn, X.~Zhai, T.~Unterthiner, M.~Dehghani, M.~Minderer, G.~Heigold, S.~Gelly, J.~Uszkoreit, and N.~Houlsby.
\newblock An image is worth 16x16 words: Transformers for image recognition at scale.
\newblock In {\em Proceedings of International Conference on Learning Representations (ICLR)}, 2021.

\bibitem{task_singular_vectors_2024}
A.~A. Gargiulo, D.~Crisostomi, M.~S. Bucarelli, S.~Scardapane, F.~Silvestri, and E.~Rodol{\`a}.
\newblock Task singular vectors: Reducing task interference in model merging.
\newblock {\em arXiv preprint arXiv:2412.00081}, 2024.

\bibitem{large_spike_slab_2012}
I.~J. Goodfellow, A.~Courville, and Y.~Bengio.
\newblock Large-scale feature learning with spike-and-slab sparse coding.
\newblock In {\em Proceedings of the 29th International Coference on International Conference on Machine Learning}, pages 1387--1394, 2012.

\bibitem{eurosat_2018}
P.~Helber, B.~Bischke, A.~Dengel, and D.~Borth.
\newblock Introducing euro{SAT}: A novel dataset and deep learning benchmark for land use and land cover classification.
\newblock In {\em Proceedings of IEEE International Geoscience and Remote Sensing Symposium}, pages 204--207, Valencia, Spain, 22--27 Jul 2018.

\bibitem{in_context_task_vectors_2023}
R.~Hendel, M.~Geva, and A.~Globerson.
\newblock In-context learning creates task vectors.
\newblock In {\em The 2023 Conference on Empirical Methods in Natural Language Processing}, 2023.

\bibitem{finding_visual_2024}
A.~Hojel, Y.~Bai, T.~Darrell, A.~Globerson, and A.~Bar.
\newblock Finding visual task vectors.
\newblock In {\em European Conference on Computer Vision}, pages 257--273. Springer, 2024.

\bibitem{badtv_2025}
C.-Y. Hsu, Y.-L. Tsai, Y.~Zhe, Y.-L. Chen, C.-H. Lin, C.-M. Yu, Y.~Zhang, C.-Y. Huang, and J.~Sakuma.
\newblock Badtv: Unveiling backdoor threats in third-party task vectors.
\newblock {\em arXiv preprint arXiv:2501.02373}, 2025.

\bibitem{multimodal_task_vector_2024}
B.~Huang, C.~Mitra, L.~Karlinsky, A.~Arbelle, T.~Darrell, and R.~Herzig.
\newblock Multimodal task vectors enable many-shot multimodal in-context learning.
\newblock {\em Advances in Neural Information Processing Systems}, 37:22124--22153, 2024.

\bibitem{task_arith_2023}
G.~Ilharco, M.~T. Ribeiro, M.~Wortsman, L.~Schmidt, H.~Hajishirzi, and A.~Farhadi.
\newblock Editing models with task arithmetic.
\newblock In {\em Proceedings of International Conference on Learning Representations (ICLR)}, Kigali, Rwanda, 1--5 May 2023. OpenReview.net.

\bibitem{spike_slab_selection_2005}
H.~ISHWARAN and J.~S. RAO.
\newblock Spike and slab variable selection: Frequentist and bayesian strategies.
\newblock {\em The Annals of Statistics}, 33(2):730--773, 2005.

\bibitem{standford_cars_2013}
J.~Krause, M.~Stark, J.~Deng, and L.~Fei-Fei.
\newblock {3D} object representations for fine-grained categorization.
\newblock In {\em IEEE/CVF International Conference on Computer Vision (ICCV) Workshop on 3D Representation and Recognition}, pages 554--561, Sydney, Australia, 1--8 Dec 2013.

\bibitem{spike_slab_posterior_2025}
S.~Kumar, P.~Sarkar, K.~Tian, and Y.~Zhu.
\newblock Spike-and-slab posterior sampling in high dimensions.
\newblock {\em arXiv preprint arXiv:2503.02798}, 2025.

\bibitem{mnist_1998}
Y.~LeCun, C.~Cortez, and C.~C.~J. Burges.
\newblock The mnist handwritten digit database, 1998.

\bibitem{generalization_analysis_2025}
H.~Li, Y.~Zhang, S.~Zhang, P.-Y. Chen, S.~Liu, and M.~Wang.
\newblock When is task vector provably effective for model editing? a generalization analysis of nonlinear transformers.
\newblock In {\em International Conference on Learning Representations}, 2025.

\bibitem{adamw_2019}
I.~Loshchilov and F.~Hutter.
\newblock Decoupled weight decay regularization.
\newblock In {\em Proceedings of International Conference on Learning Representations (ICLR)}, New Orleans, LA, USA, 6--9 may 2019. OpenReview.net.

\bibitem{vlm_task_vector_2025}
G.~Luo, T.~Darrell, and A.~Bar.
\newblock Vision-language models create cross-modal task representations.
\newblock In {\em ICML}, 2025.

\bibitem{comparing_spike_slab_2011}
G.~Malsiner-Walli and H.~Wagner.
\newblock Comparing spike and slab priors for bayesian variable selection.
\newblock {\em Austrian Journal of Statistics}, 40(4):241--264, 2011.

\bibitem{bayesian_variable_linear_1988}
T.~MITCHELL, J.~BEAUCHAMP, J.~BERGER, and C.~MALLOWS.
\newblock Bayesian variable selection in linear regression. comments.
\newblock {\em Journal of the American Statistical Association}, 83(404):1023--1036, 1988.

\bibitem{spike_slab_discovery_2021}
R.~Nayek, R.~Fuentes, K.~Worden, and E.~J. Cross.
\newblock On spike-and-slab priors for bayesian equation discovery of nonlinear dynamical systems via sparse linear regression.
\newblock {\em Mechanical Systems and Signal Processing}, 161:107986, 2021.

\bibitem{svhn_2011}
Y.~Netzer, T.~Wang, A.~Coates, A.~Bissacco, B.~Wu, and A.~Y. Ng.
\newblock Reading digits in natural images with unsupervised feature learning.
\newblock In {\em Neural Information Processing Systems (NeurIPS) Workshop on Deep Learning and Unsupervised Feature Learning}, Granada, Spain, 12--17 Dec 2011.

\bibitem{tangent_space_2023}
G.~Ortiz-Jim{\'{e}}nez, A.~Favero, and P.~Frossard.
\newblock Task arithmetic in the tangent space: Improved editing of pre-trained models.
\newblock In {\em Advances in Neural Information Processing Systems (NeurIPS)}, volume~36, pages 66727--66754, New Orleans, LA, USA, 10--16 Dec 2023. Curran Associates, Inc.

\bibitem{clip_2021}
A.~Radford, J.~W. Kim, C.~Hallacy, A.~Ramesh, G.~Goh, S.~Agarwal, G.~Sastry, A.~Askell, P.~Mishkin, J.~Clark, G.~Krueger, and I.~Sutskever.
\newblock Learning transferable visual models from natural language supervision.
\newblock In {\em Proceedings of International Conference on Machine Learning (ICML)}, volume 139, pages 8748--8763. Proceedings of Machine Learning Research (PMLR), 18--24 Jul 2021.

\bibitem{gpt2_2019}
A.~Radford, J.~Wu, R.~Child, D.~Luan, D.~Amodei, and I.~Sutskever.
\newblock Language models are unsupervised multitask learners.
\newblock {\em OpenAI blog}, 2019.

\bibitem{t5_2020}
C.~Raffel, N.~Shazeer, A.~Roberts, K.~Lee, S.~Narang, M.~Matena, Y.~Zhou, W.~Li, and P.~J. Liu.
\newblock Exploring the limits of transfer learning with a unified text-to-text transformer.
\newblock {\em Journal of Machine Learning Research}, 21(140):1--67, 2020.

\bibitem{spike_slab_logit_2020}
K.~Ray, B.~Szab{\'o}, and G.~Clara.
\newblock Spike and slab variational bayes for high dimensional logistic regression.
\newblock {\em Advances in Neural Information Processing Systems}, 33:14423--14434, 2020.

\bibitem{rethinking_VI_2022}
T.~Reichelt, L.~Ong, and T.~Rainforth.
\newblock Rethinking variational inference for probabilistic programs with stochastic support.
\newblock {\em Advances in Neural Information Processing Systems}, 35:15160--15175, 2022.

\bibitem{dynamic_spike_slab_2021}
V.~Rockova and K.~McAlinn.
\newblock Dynamic variable selection with spike-and-slab process priors.
\newblock {\em Bayesian Analysis}, 16(1):233--269, 2021.

\bibitem{spike_slab_structured_2012}
F.~Scheipl, L.~Fahrmeir, and T.~Kneib.
\newblock Spike-and-slab priors for function selection in structured additive regression models.
\newblock {\em Journal of the American Statistical Association}, 107(500):1518--1532, 2012.

\bibitem{gtsrb_2011}
J.~Stallkamp, M.~Schlipsing, J.~Salmen, and C.~Igel.
\newblock The german traffic sign recognition benchmark: A multi-class classification competition.
\newblock In {\em Proceedings of International Joint Conference on Neural Networks (IJCNN)}, pages 1453--1460, San Jose, CA, USA, 31 Jul--5 Aug 2011.

\bibitem{spike_slab_vi_2011}
M.~Titsias and M.~L{\'a}zaro-Gredilla.
\newblock Spike and slab variational inference for multi-task and multiple kernel learning.
\newblock {\em Advances in neural information processing systems}, 24, 2011.

\bibitem{function_vectors_llm_2024}
E.~Todd, M.~Li, A.~S. Sharma, A.~Mueller, B.~C. Wallace, and D.~Bau.
\newblock Function vectors in large language models.
\newblock In {\em The Twelfth International Conference on Learning Representations}, 2024.

\bibitem{knowledge_editing_survey_2024}
S.~Wang, Y.~Zhu, H.~Liu, Z.~Zheng, C.~Chen, and J.~Li.
\newblock Knowledge editing for large language models: A survey.
\newblock {\em ACM Computing Surveys}, 57(3):1--37, 2024.

\bibitem{sun397_2016}
J.~Xiao, K.~A. Ehinger, J.~Hays, A.~Torralba, and A.~Oliva.
\newblock Sun database: Exploring a large collection of scene categories.
\newblock {\em Interational Journal of Computer Vision (IJCV)}, 119(1):3--22, 2016.

\bibitem{knowledge_conflicts_2024}
R.~Xu, Z.~Qi, Z.~Guo, C.~Wang, H.~Wang, Y.~Zhang, and W.~Xu.
\newblock Knowledge conflicts for llms: A survey.
\newblock In {\em Proceedings of the 2024 Conference on Empirical Methods in Natural Language Processing}, pages 8541--8565, 2024.

\bibitem{model_merging_2024}
E.~Yang, L.~Shen, G.~Guo, X.~Wang, X.~Cao, J.~Zhang, and D.~Tao.
\newblock Model merging in llms, mllms, and beyond: Methods, theories, applications and opportunities.
\newblock {\em arXiv preprint arXiv:2408.07666}, 2024.

\bibitem{knowledge_composition_2024}
F.~Z. Zhang, P.~Albert, C.~Rodriguez-Opazo, A.~van~den Hengel, and E.~Abbasnejad.
\newblock Knowledge composition using task vectors with learned anisotropic scaling.
\newblock {\em Advances in Neural Information Processing Systems}, 37:67319--67354, 2024.

\bibitem{composing_arithmetic_2023}
J.~Zhang, J.~Liu, J.~He, et~al.
\newblock Composing parameter-efficient modules with arithmetic operation.
\newblock {\em Advances in Neural Information Processing Systems}, 36:12589--12610, 2023.

\bibitem{comprehensive_knowledge_editing_2024}
N.~Zhang, Y.~Yao, B.~Tian, P.~Wang, S.~Deng, M.~Wang, Z.~Xi, S.~Mao, J.~Zhang, Y.~Ni, et~al.
\newblock A comprehensive study of knowledge editing for large language models.
\newblock {\em arXiv preprint arXiv:2401.01286}, 2024.

\bibitem{dynamic_grouping_2025}
P.~Zhang, R.~Zhang, and Z.~Nie.
\newblock Dynamic task vector grouping for efficient multi-task prompt tuning.
\newblock {\em arXiv preprint arXiv:2503.18063}, 2025.

\end{thebibliography}
